\documentclass[a4paper,fleqn]{cas-dc}

\usepackage[numbers]{natbib}
\usepackage[nodots]{numcompress}
\usepackage{algorithm}
\usepackage{algorithmic}
\usepackage{subfigure}
\usepackage{hyperref}
\usepackage{threeparttable}
\usepackage{multicol}
\usepackage{multirow, bm}

\def\tsc#1{\csdef{#1}{\textsc{\lowercase{#1}}\xspace}}
\tsc{WGM}
\tsc{QE}

\begin{document}
\let\WriteBookmarks\relax
\def\floatpagepagefraction{1}
\def\textpagefraction{.001}

\shorttitle{A Surrogate-Assisted Controller for Expensive Evolutionary Reinforcement Learning}    

\shortauthors{Yuxing Wang et al.}  

\title [mode = title]{A Surrogate-Assisted Controller for Expensive Evolutionary Reinforcement Learning}  


%

\author[1]{Yuxing Wang}




\affiliation[1]{organization={Shenzhen International Graduate School},
            addressline={Tsinghua University}, 
            city={Shenzhen},
            postcode={518055}, 
            country={China},
            }
            
\author[1]{Tiantian Zhang}

\author[1]{Yongzhe Chang}
\author[2]{Bin Liang}
\affiliation[2]{organization={Department of Automation},
            addressline={Tsinghua University}, 
            city={Beijing},
            postcode={100084}, 
            country={China}}

\author[1]{Xueqian Wang}

\author[1]{Bo Yuan}
\cormark[1]
\cortext[1]{Corresponding author at: The Center for Artificial Intelligence and Robotics, Shenzhen International Graduate School, Tsinghua University, Shenzhen 518055, China. E-mail address:boyuan@ieee.org.}

\begin{abstract}
The integration of Reinforcement Learning (RL) and Evolutionary Algorithms (EAs) aims at simultaneously exploiting the sample efficiency as well as the diversity and robustness of the two paradigms. Recently, hybrid learning frameworks based on this principle have achieved great success in various challenging tasks. However, in these methods, policies from the genetic population are evaluated via interactions with the real environments, limiting their applicability when such interactions are prohibitively costly. In this work, we propose Surrogate-assisted Controller (SC), a novel and efficient module that can be integrated into existing frameworks to alleviate the burden of EAs by partially replacing the expensive fitness evaluation. The key challenge in applying this module is to prevent the optimization process from being misled by the possible false minima introduced by the surrogate. To address this issue, we present two strategies for SC to control the workflow of hybrid frameworks. Experiments on six continuous control tasks from the OpenAI Gym platform show that SC can not only significantly reduce the cost of interacting with the environment, but also boost the performance of the original hybrid frameworks with collaborative learning and evolutionary processes.
\end{abstract}

\maketitle

\section{Introduction}\label{intro}
Reinforcement Learning (RL) has demonstrated promising achievements in various domains, ranging from Atari games \cite{mnih2015human}, GO \cite{silver2016mastering}, to robot control tasks \cite{lillicrap2015continuous}. Among these successes, Deep Learning (DL) techniques \cite{goodfellow2016deep} such as Deep Neural Networks (DNNs) have been widely used for decision-making \cite{sutton2018reinforcement,wilson2018evolving}. The combination of RL with DL is generally called Deep Reinforcement Learning (DRL). However, recent studies show that DRL suffers from premature convergence to local optima in the training process, and the RL agents are highly sensitive to hyperparameter settings, implementation details and uncertainties of the environmental dynamics \cite{ilyas2018closer}, preventing agents from learning stable policies.

In the meantime, black-box optimization techniques such as Evolutionary Algorithms (EAs) \cite{yao1999evolving, drugan2019reinforcement} have shown competitive results compared to DRL algorithms \cite{salimans2017evolution, yang2021parallel}. Firstly, the population-based mechanism makes EAs explore the parameter space better than DRL. Secondly, since EAs only consider the total returns across the entire episode, they are indifferent to the issue of sparse reward and robust to environmental noise \cite{salimans2017evolution}. However, EAs suffer from low sample efficiency \cite{khadka2018evolution}, due to their inherent black-box properties, and they often do not make full use of the feedback signals and historical data from the environment.

\begin{figure}[t]
\centering
\includegraphics[width=\columnwidth]{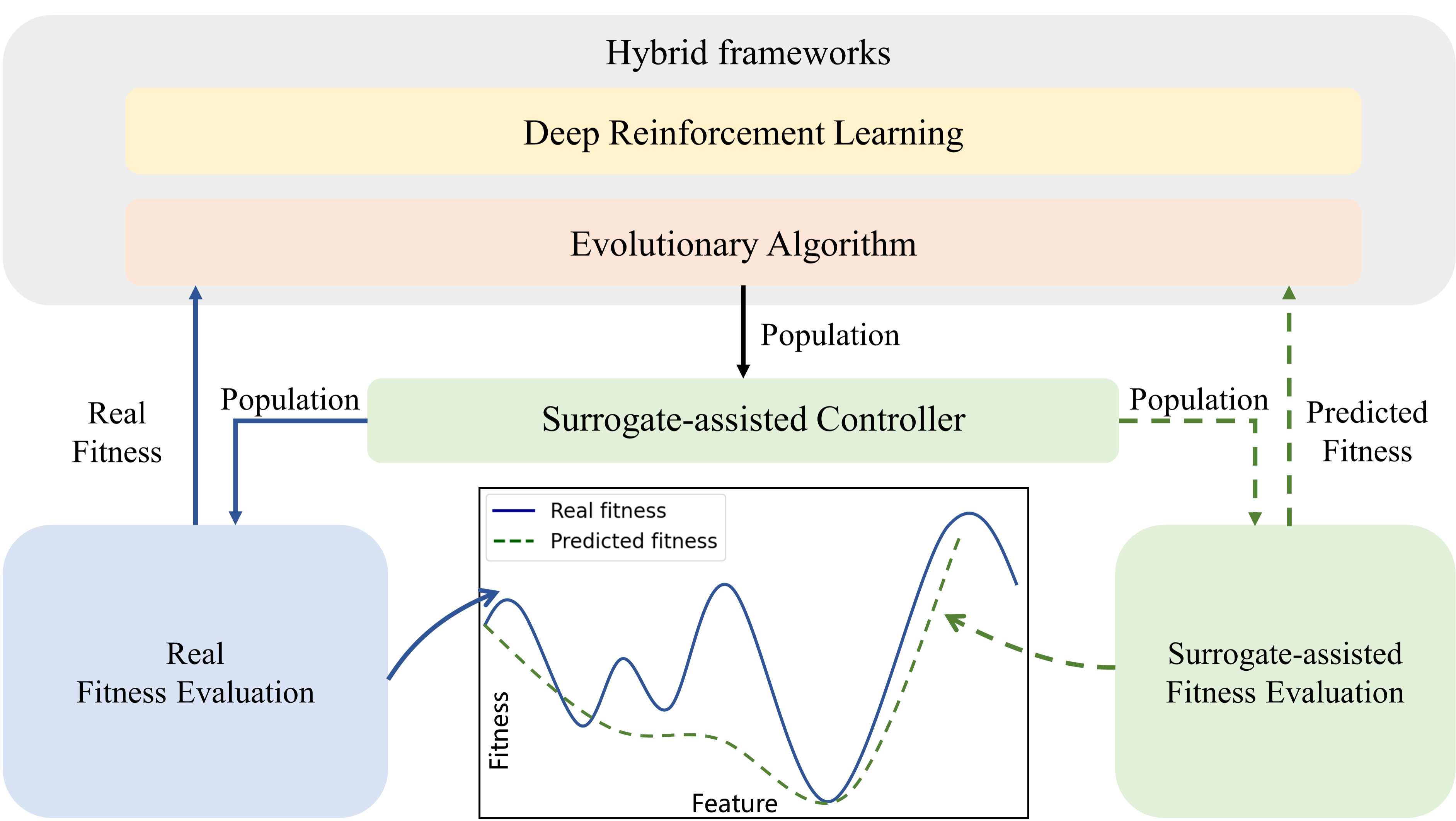} 
\caption{An overview of Surrogate-assisted Controller (SC). The SC switches between the real fitness function and the approximated fitness function constructed by the surrogate model to efficiently evaluate the genetic population in hybrid frameworks, while boosting the performance of the original hybrid frameworks. As shown at the bottom, in some cases, inaccurate surrogates may bring extra benefits: the predicted fitness function (dashed curve) has large deviations from the real one (solid curve) but nevertheless features the same global minimum and constructs a new fitness landscape that is more friendly to evolutionary search.}
\label{sc}
\end{figure}

An emergent research direction is dedicated to exploiting the benefits of both solutions following the theory of evolution \cite{weber2003evolution}, where the learning of individuals can increase the evolutionary advantage of species, which can subsequently make the population learn faster. Recently proposed Evolutionary Reinforcement Learning (ERL) \cite{khadka2018evolution}, Proximal Distilled ERL (PDERL) \cite{bodnar2020proximal} and other hybrid frameworks based on EA and off-policy DRL \cite{pourchot2018cem,marchesini2020genetic,yang2022evolutionary,lu2021recruitment} have demonstrated encouraging progresses on single-objective hard-exploration tasks with large continuous state and action spaces \cite{brockman2016openai}, outperforming pure DRL and EA. Particularly, one common feature of these approaches is that a diverse genetic population of policies is generated by EA to drive the exploration, while an off-policy actor-critic algorithm thoroughly exploits the population's environmental experience to search for high-performing policies.

However, there is a notable issue with these hybrid frameworks: the evaluation of the genetic population requires all individuals to be presented to the environment during the training process. Although population-based interactions can provide diverse historical experiences, they usually require a large amount of computational time and can be prohibitively expensive \cite{schneider2000neural,jin2001managing}. For instance, in robotic scenarios, frequent evaluations may lead to massive resources consumption or even equipment damage. These unfavorable characteristics greatly limit the applicability of existing hybrid frameworks in complex simulated circumstances or non-trivial real-world scenarios.
 
A possible solution is to introduce the mechanism of surrogate. In typical Surrogate-assisted Evolutionary Optimization (SEO), surrogates, also known as meta-models, are trained to partially replace the expensive fitness functions \cite{tong2021surrogate,pan2021efficient}. Surrogate models such as Kriging \cite{dong2021kriging}, polynomials \cite{keane2008engineering} and neural networks have been successfully applied to domains including constrained optimization \cite{audet2000surrogate} and multi-objective optimization \cite{loshchilov2010mono, wang2020adaptive}. Nevertheless, in the standard RL context, due to the large uncertainties of genotype-phenotype-fitness mapping \cite{stork2019surrogate}, conventional methods face great challenges of high computational cost and modeling complexity (Section \ref{smbo}) with only very limited studies conducted on simple discrete control tasks. 

In this paper, we design a generic and effective module, called Surrogate-assisted Controller (SC), that can be conveniently combined with existing hybrid RL frameworks to improve their practicability. As shown in Figure \ref{sc}, SC employs an approximated fitness function together with the real fitness function to help evaluate the genetic population. Its sample-efficient surrogate model can be naturally implemented in the existing hybrid RL frameworks, making full use of the diverse experiences to evaluate the fitness of individuals without environmental interactions. Note that the primary criterion for applying the surrogate model is the introduced prediction error \cite{ratle1999optimal}. SC mitigates this risk using two management strategies with an elite protection mechanism (Section \ref{mana}), which can strategically schedule the real and the surrogate-assisted evaluation as well as effectively prevent the dramatic fluctuation of performance and the spread of detrimental information from individuals with inaccurate fitness values. Our empirical studies in Section \ref{experiment} show that SC can not only successfully transfers the computational burden of real fitness evaluations to the efficient surrogate model and boost the performance of original hybrid approaches, but also stabilize the interactions between the RL agent and the genetic population. 

The major contributions of our work are summarized as follows:

\begin{itemize}
\item A novel module named Surrogate-assisted Controller (SC) is proposed that can be easily integrated into existing hybrid RL frameworks to significantly relieve the computational cost of interacting with the real environment during the optimization process.

\item Two effective management strategies are presented for SC to control the workflow of the hybrid frameworks, which can strategically schedule the surrogate-assisted evaluation and the real fitness evaluation.

\item We combine SC with ERL and PDERL, creating two new frameworks named SERL and SPDERL, respectively, to highlight its principle and flexibility. Comprehensive experimental studies on Mujoco benchmarks show that SC can not only effectively bring better sample-efficiency to the original hybrid frameworks, but also stabilize the learning and evolution processes with superior performance. 
\end{itemize}

The remainder of this paper is organized as follows. Section \ref{related_work} introduces the related work on hybrid frameworks and surrogate-assisted methods for solving RL problems. The problem definition is specified in Section \ref{prelimi} and the details of SC and its components are presented in Section \ref{Methodology}. In Section \ref{experiment}, the numerical validation and in-depth analyses are conducted and Section \ref{conclusion} concludes our work with some discussions on future research directions.

\section{Related work}\label{related_work}
\subsection{Combining EA and off-policy DRL}
As an alternative solution for RL problems \cite{salimans2017evolution}, EAs have also been combined with DRL to leverage the benefits of both solutions. The first hybrid framework ERL \cite{khadka2018evolution} combines an off-policy DRL agent based on DDPG \cite{lillicrap2015continuous} with a population evolved by the Genetic Algorithm (GA) \cite{whitley1994genetic}. In each generation, individuals are evaluated over a few episodes and the fitness is given by averaging the total rewards. Based on this information, policy optimization is then conducted over the parameter space by genetic operators. Meanwhile, the RL agent is trained on the diverse experiences produced by the population. In the periodical synchronization step, it is injected into the population and this bi-directional interaction controls the information flow between the RL agent and the genetic population. To further extend the ERL framework, PDERL \cite{bodnar2020proximal} uses the distillation crossover operator to alleviate the catastrophic forgetting caused by the recombination of neural networks. In addition, various evolutionary methods \cite{pourchot2018cem,marchesini2020genetic,suri2020maximum} have also been combined with other DRL frameworks such as TD3 \cite{fujimoto2018addressing} and SAC \cite{haarnoja2018soft} to further leverage the advantages of both gradient-based and gradient-free methods. In our paper, we mainly focus on the paradigm of ERL and PDERL to show the effectiveness of our proposed methods.

\subsection{Surrogate-assisted methods for RL problems} \label{smbo}
Surrogate-assisted methods have been widely investigated to reduce unnecessary expensive evaluations \cite{jin2011surrogate}. Recently, a few studies on using the surrogate model to enhance RL algorithms or EAs in the context of sequential decision-making tasks have been conducted and can be generally divided into two categories.

The first category focuses on learning a model of the environment. Many studies \cite{cao2022model,ha2018recurrent,deisenroth2011pilco} try to construct a transition model, which is trained in a standard supervised manner using a large amount of historical data, to imitate at least some aspects of the system’s physical dynamics. Then, the fixed surrogate model is embedded into RL algorithms to help learn a control strategy from the experiences obtained by continuous interactions with the surrogate model. However, for domains with high noise or when the availability of the historical data is limited, this kind of methods may no longer be effective.

The other category does not require modeling the environment and mainly focuses on evaluation. For example, Kriging can be employed to directly map the relationships between neural networks (policies) and their fitness. Typically, genotypic distances between neural networks are needed by the Kriging model \cite{stork2019improving}. However, the large-scale settings and complex problems may make the computation of these distances practically impossible. Although approximate distances such as the phenotypic distance \cite{stork2019surrogate} can offer some help, the high dimensionality of the inputs can result in an increase of computational costs, and the parameter settings for Kriging remain a challenge for input vectors of different sizes \cite{stork2019surrogate}.

For instance, Evolutionary Surrogate-assisted Prescription (ESP) \cite{francon2020effective} incorporates a surrogate model into the EA. Given a set of input states $\mathscr{S}$, the policy neural network takes each $s \in \mathscr{S}$ as the input and outputs the action $a$. The surrogate model of ESP, represented by a random forest or a deep neural network, is used to predict the outcome of each state-action pair $(s, a)$, and the fitness of each policy is given by averaging the outcomes over $\mathscr{S}$. In each generation, the surrogate is first trained on the historical data by minimizing the Mean Square Error (MSE) loss between the real and the predicted outcomes. Subsequently, individuals (called $prescriptors$) in the population are evolved with the trained surrogate. Finally, selected elites are presented to the real environment to generate new training data for the surrogate. Unfortunately, for challenging environments with large continuous state and action spaces or with rich feedback signals, the lack of gradient information makes EAs suffer from brittle convergence \cite{salimans2017evolution, khadka2018evolution}. Furthermore, the 
update of the policy neural networks in ESP is purely based on predicted fitness, which increases the risk of misleading the evolutionary optimization in the wrong direction under complex circumstances. Consequently, the frequency of applying the surrogate also needs to be managed properly to ensure the stability of the training process.

In general, apart from limited successes on simple discrete control tasks such as Cart-Pole \cite{brockman2016openai}, conventional surrogate-assisted EAs \cite{stork2019surrogate,stork2019improving,francon2020effective} still face significant challenges. To explore the potential of surrogate-assisted methods in continuous and complex RL contexts, in this work, we extend the surrogate model to hybrid RL frameworks with the objective to reduce the  cost of evaluations while making full use of the efficiency of the RL agent and the exploration capability of the EA.
 
\section{Preliminaries}\label{prelimi}
In DRL, each problem is modeled as a Markov Decision Process (MDP), which can be specified by a 5-tuple $\left \langle \mathscr{S}, \mathscr{A}, \mathscr{P},r, \gamma  \right \rangle$. With the state space $\mathscr{S}$ and the action space $\mathscr{A}$, $\mathscr{P}:\mathscr{S} \times \mathscr{S} \times \mathscr{A} \to [0, 1]$ is the transition function of the environment; $r(s,a):\mathscr{S} \times \mathscr{A} \to \mathbb{R}$ is the reward function, and the discount factor $\gamma \in (0,1]$ specifies the degree to which rewards are discounted over time. 

\begin{equation}
Q(s,a|\theta^{Q})=\mathbb{E}\Bigg[{\sum_{i=0}^{\infty}\gamma ^{i}r_{t+i+1}\big|s_{t}=s,a_{t}=a}\Bigg] 
\label{eq:qvalue}
\end{equation}

The actor-critic architecture based on the policy gradient approach is widely used in DRL \cite{sutton2018reinforcement}. The critic network $Q(s,a|\theta^{Q})$, as shown in Eq.(\ref{eq:qvalue}), is used to estimate the expectation of the discounted accumulative rewards of the state-action pair $(s,a)$. According to the Bellman equation, its recursive expression is:
\begin{equation}
 Q(s,a|\theta^{Q})=\mathbb{E}\big[r(s,a)+\gamma Q(s^\prime,a^\prime|\theta^{Q})\big]
\label{eq:critic}
\end{equation}
where $s^\prime$ and $a^\prime$ represent the next state and action, respectively. In each iteration, transitions with the batch size of $N$ are sampled randomly from the experience replay buffer to update the parameters of $Q(s,a|\theta^{Q})$ by minimizing the Temporal Difference (TD) loss based on the standard back-propagation: 

\begin{equation}
    \mathscr{L}_{Q(s,a|\theta^{Q})}=\frac{1}{N}\sum_{i}\big(y_i-Q(s_{i},a_{i}|\theta^{Q})\big)^{2} 
    \label{eq:loss1}
\end{equation}

\begin{equation}
    y_{i}=r(s_i,a_i)+\gamma Q\big(s^\prime,\mu(s^\prime|\theta^{\mu})|\theta^{Q}\big)
    \label{eq:loss2}
\end{equation}

\begin{equation}
    \nabla_{\theta^{\mu}}J\approx\frac{1}{N}\sum_{i}Q\big(s_{i},\mu(s_{i}|\theta^{\mu})\big)\nabla_{\theta^{\mu}}\mu(s_{i}|\theta^{\mu})
    \label{eq:loss3}
\end{equation}

Then, the critic is used to assist the training of the actor network $\mu(s|\theta^\mu)$ via policy gradient, according to Eq.(\ref{eq:loss3}). With the back-propagation method, the parameter $\theta^\mu$ of the actor network is updated in the direction towards maximizing the $Q$ value. 

\section{Methodology}\label{Methodology}
\begin{figure}[t]
\centering
\includegraphics[width=\columnwidth]{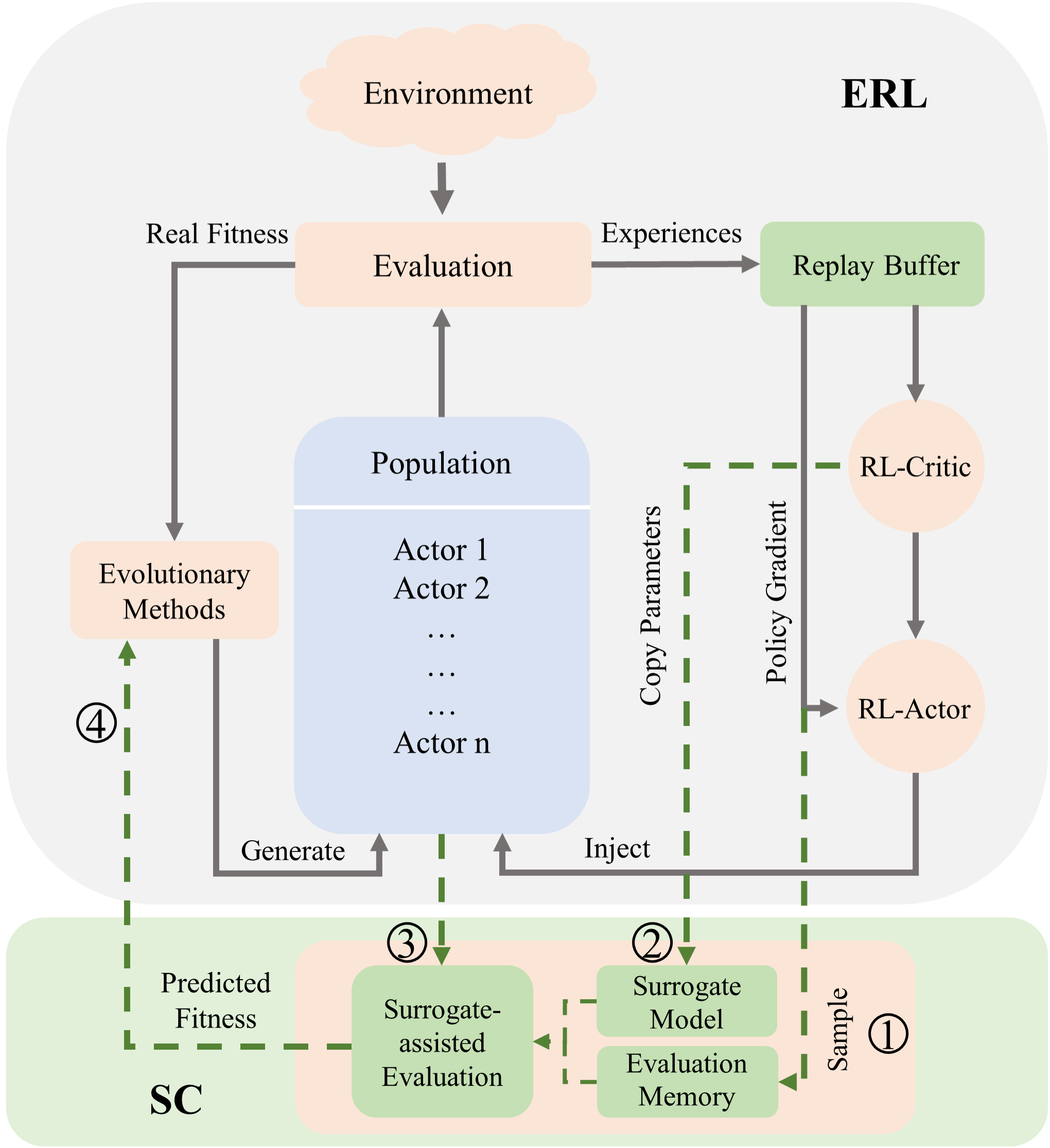} 
\caption{Integration of SC and ERL. The ERL framework is shown in the light gray box and SC is shown in the light green box. To evaluate the population without environment interactions, the recent state information is sampled from the replay buffer as the evaluation memory. After that, the RL-Critic plays the role of a surrogate model to evaluate actors based on the historical data. Finally, the predicted population fitness is consumed by evolutionary methods.}
\label{fig2}
\end{figure}

In this section, we present the Surrogate-assisted Controller (SC) with two management strategies and introduce how to incorporate it into the hybrid framework. Figure \ref{fig2} shows an example of combining SC with ERL, where the RL-Critic exploits the diverse experiences collected by the genetic population to simultaneously train the RL-Actor and assist the evaluation, referred to as Surrogate-assisted Evolutionary Reinforcement Learning (SERL).  In practice, SC includes three components: critic-based surrogate model, management strategies and evaluation memory.

\begin{algorithm}[t]
\caption{Surrogate-assisted Evaluation}
\label{alg:algorithm1}
\textbf{Input}: Policy set of population $\mu_{pop}=\{\mu_{1},\mu_{2},...,\mu_{n}\}$; 
\newline \hspace*{0.95cm} The surrogate model $Q(s,a|\theta^{Q})$;
\newline \hspace*{0.95cm} Replay buffer $\mathcal{R}$ for RL agent training.\\
\textbf{Output}: Population fitness $F$
\begin{algorithmic}[1] 
\STATE Sample $k$ latest states from $\mathcal{R}$ to the evaluation memory $\mathcal{R}_{eva}$ as the evaluation samples;
\FOR{$i=1$ to $n$}
\STATE Initialize the fitness of $\mu_i$: $f_{i}=0$;
\FOR{$j=1$ to $k$}
\STATE $f_{i}=f_{i}+Q\big(s_{j},\mu_{i}(s_{j}|\theta^{\mu_{i}})|\theta^{Q}\big)/k$;
\ENDFOR
\ENDFOR 
\STATE \textbf{return} Population fitness $F=\{f_{1},f_{2},...,f_{n}\}$.
\end{algorithmic}
\end{algorithm}

\subsection{Critic-based surrogate model}\label{critic-surrogate}
Using the real fitness function can be time-consuming or even dangerous for expensive problems. A surrogate, which is a predictive model, can be used to assist the evaluation. Note that, in the domain of Evolutionary Computation (EC), the fitness of an individual can be determined in many forms, even as a non-markovian definition. In our work, we consider the real fitness value as the sum of the reward over the episode, following the typical setting of RL paradigm. 

\begin{figure}
\centering
\includegraphics[width=0.49\columnwidth]{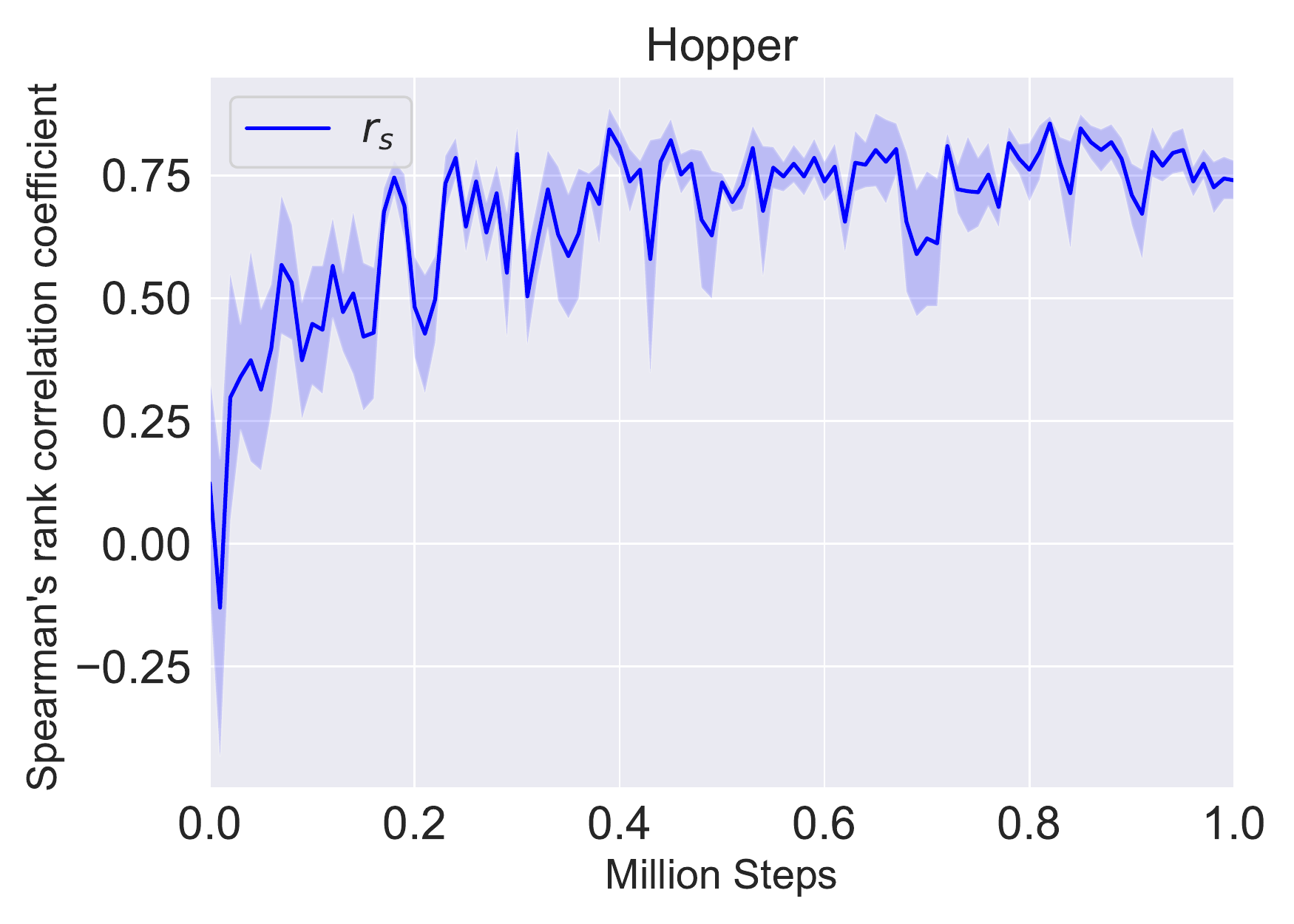} 
\includegraphics[width=0.49\columnwidth]{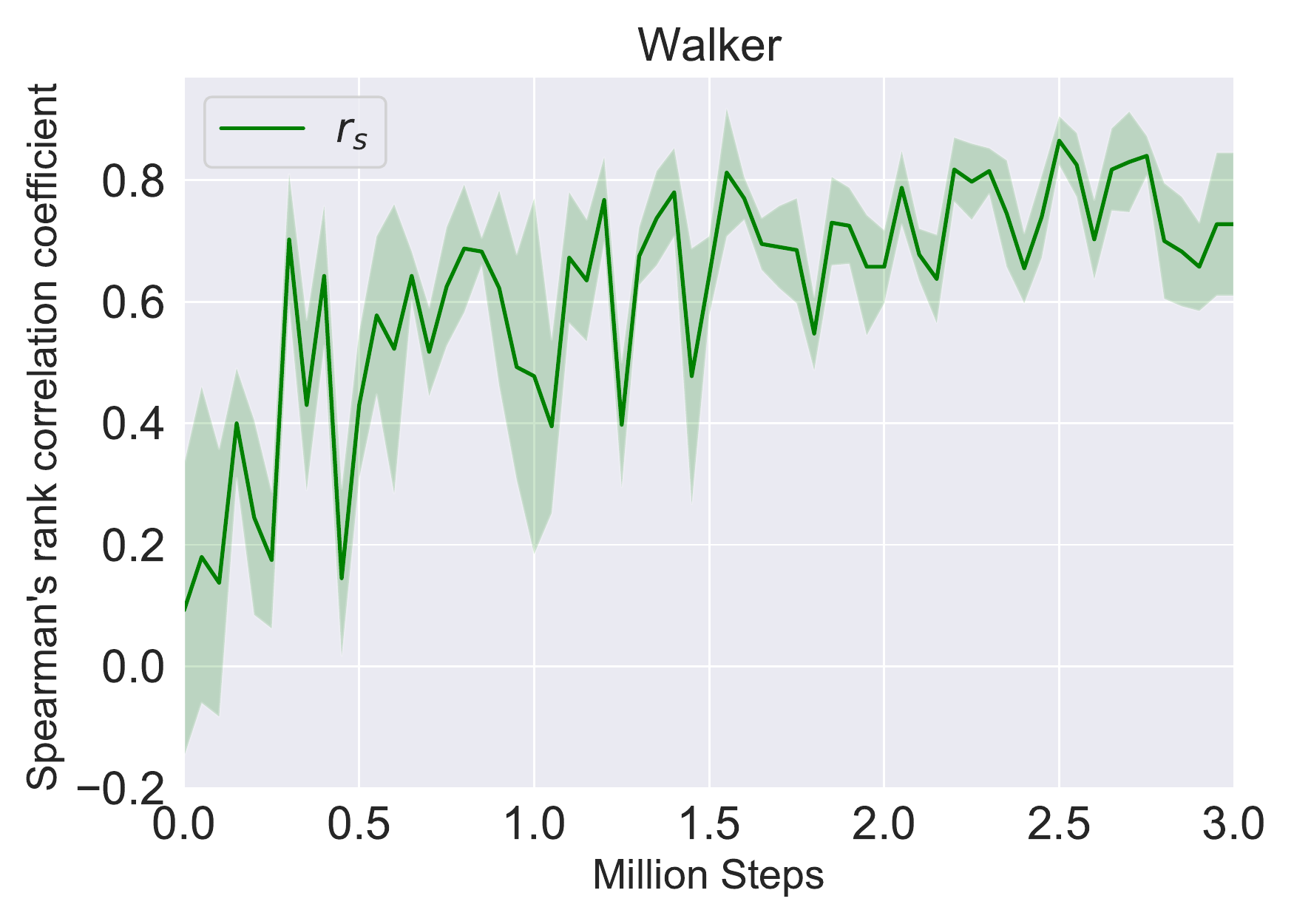}
\caption{The preliminary experiment on the surrogate's approximation accuracy over two environments: Hopper (left, trained for 1 million steps) and Walker (right, trained for 3 million steps). The curves of approximation accuracy clearly show that the accuracy of the surrogate tends to increase during the training process.}
\label{fig3}
\end{figure}

However, the genotype-phenotype-fitness mapping of sequential decision-making problems is often difficult to learn, but with the definition of fitness mentioned above, it is straightforward for the surrogate model to estimate the outcome of a state-action pair $(s,a)$, as shown in Eq.(\ref{eq:qvalue}). In ESP \cite{francon2020effective}, individuals are evaluated by an extra dedicated surrogate model. However, in hybrid frameworks, the critic module $Q(s,a|\theta^{Q})$ of the off-policy RL agent can be naturally used as a surrogate. The actor module $\mu(s|\theta^\mu)$ represented by a fully connected policy neural network takes a state vector $s$ as the input and decides what action vector $a$ to perform. Then, the concatenated vector $(s,a)$ is evaluated by the critic-based surrogate model. Thus, with the evaluation memory (described in Section \ref{evalmem}) that contains the information of the $k$ latest state vectors drawn from the replay buffer, the fitness value of an actor in the genetic population can be obtained by averaging its predicted $Q$ values over all states. Then we have:

\begin{equation}
f= {\frac{1}{k} \textstyle \sum_{j=1}^{k}Q\big(s_{j},\mu(s_{j}|\theta^{\mu})|\theta^{Q}\big)}
\label{eq:fitness}
\end{equation}

The complete process of the surrogate-assisted evaluation is presented in Algorithm \ref{alg:algorithm1}, and its intrinsic motivation is to evaluate a policy whether it is powerful and robust enough to perform as much as high-quality actions when facing diverse states. Furthermore, a prominent feature is that there is no extra cost involved in training the surrogate model, as it is part of the standard training procedure of the RL. Different from conventional actor-critic methods such as Asynchronous Advantage Actor-Critic (A3C) \cite{mnih2016asynchronous}, where the critic is only used to guide the improvement of policies, the critic-based surrogate model is able to simultaneously train the RL-Actor via policy gradient and provide relatively accurate fitness estimations for individuals with various improvements in the estimation of $Q$ values \cite{fujimoto2018addressing,van2016deep}.

The approximation quality of the surrogate is measured by the Spearman's rank correlation coefficient $r_{s}\in[-1,1]$ between the real and the predicted fitness values, where $n$ is the number of data points (population size) and $d$ represents the difference between the two ranks:

\begin{equation}
r_{s} =1 - \frac{6 {\textstyle \sum_{i=1}^{n}d_{i}^{2}}}{n(n^{2}-1)} 
\label{eq:sr_rate}
\end{equation}

A preliminary experiment on two standard continuous control tasks from MuJoCo \cite{todorov2012mujoco} is conducted to show how the approximation accuracy of the critic-based surrogate changes during ERL's training process. In each generation, we apply the surrogate-assisted evaluation to the newly generated population with $k=50,000$ in the first place. After evaluating the population in the real environment, $r_{s}$ is calculated to report the current approximation accuracy of the surrogate. The result in Figure \ref{fig3} is reported over 6 runs, it suggests that the approximation accuracy increase during the training process. Although at the beginning of training, the surrogate has relatively low approximation accuracy, previous studies \cite{jin2011surrogate,francon2020effective} suggest that this kind of uncertainty may not cause negative effects on evolutionary search. Instead, it may further push the population to explore the fitness landscape. Moreover, the training data generated by these candidates can subsequently help improve the approximation accuracy of the surrogate. 

\begin{figure*}[t]
\centering
\includegraphics[width=\textwidth]{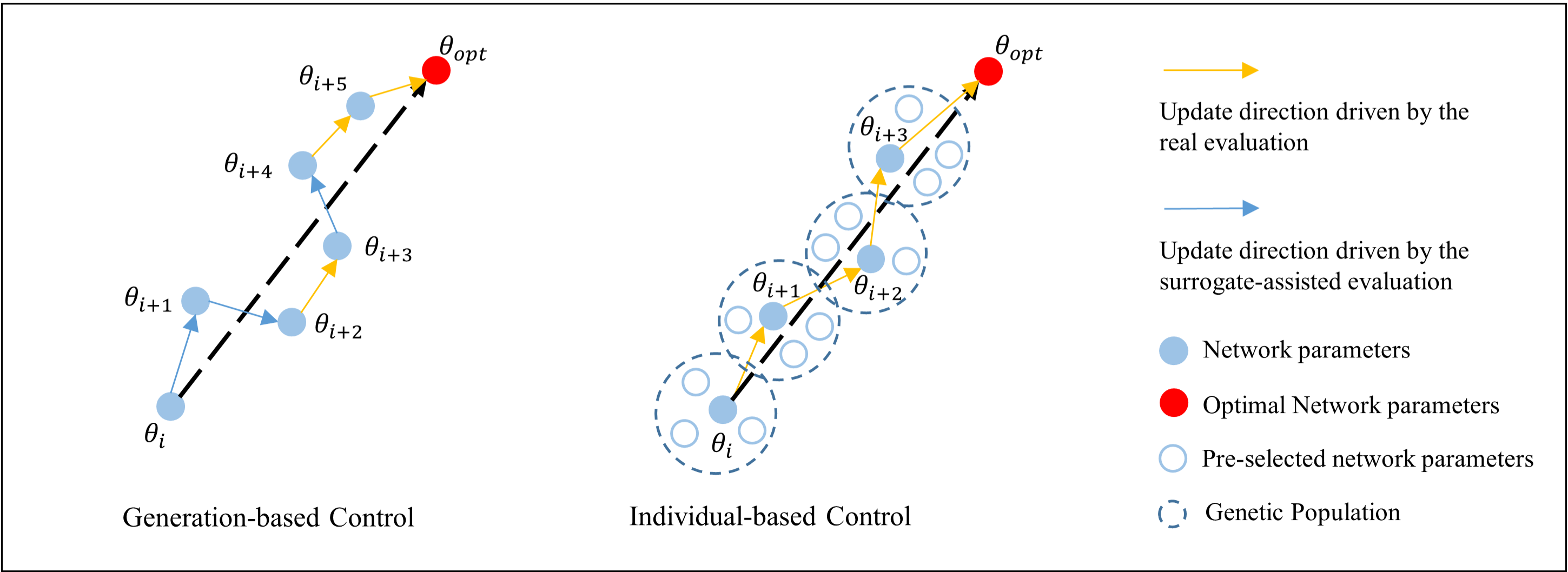}
\caption{Two kinds of management strategies: the Generation-based Control (GC) and the Individual-based Control (IC). For GC, the parameter update direction is partially based on real fitness information, while the update direction is totally provided by the real fitness of the pre-selected candidates in IC.}
\label{gen-indi}
\end{figure*}

\subsection{Management strategies}\label{mana}
Although the surrogate model can provide an approximately accurate fitness estimation, using predicted fitness to assist the evolutionary operations throughout the optimization process may easily introduce false minima, leading to a drop in performance or an increase in the computational cost of evaluating low-performing policies. As a result, surrogates should be used along with real fitness functions and we consider the following two methods: generation-based and individual-based strategies. 

\begin{algorithm}
\caption{Generation-based control}
\label{alg:gen_based_control}
\textbf{Input}: Policy set of population $\mu_{pop}=\{\mu_{1},\mu_{2},...,\mu_{n}\}$; 
\newline \hspace*{0.95cm} The surrogate model $Q_{rl}$;
\newline \hspace*{0.95cm} Replay buffer $\mathcal{R}$ for RL agent training;
\newline \hspace*{0.95cm} Control factor $\omega$; 
\newline \hspace*{0.95cm} Random number generator $G_{rand}\in(0,1]$.\\ 
\textbf{Output}: New population $\mu_{pop^{*}}$ 
\begin{algorithmic}[1] 
\IF{$G_{rand}>\omega$}
\STATE $F_{real}$ = Evaluation($\mu_{pop}$)
\STATE Copy the elite actor
\STATE $\mu_{pop^{*}}$ = Evolutionary methods($F_{real}$, $\mu_{pop}$)
\ELSE 
\STATE $F_{pre}$ = Surrogate-assisted evaluation($\mu_{pop}$, $Q_{rl}$, $\mathcal{R}$)
\STATE $\mu_{pop^{*}}$ = Evolutionary methods($F_{pre}$, $\mu_{pop}$)
\STATE Maintain the recorded elite actor in $\mu_{pop^{*}}$
\ENDIF
\STATE \textbf{return} New population $\mu_{pop^{*}}$ 
\end{algorithmic}
\end{algorithm}

\subsubsection{Generation-based control}
In the generation-based control, a fixed hyperparameter $\omega \in [0,1)$ is used to indicate the probability of using the surrogate-assisted evaluation. In this setting, the evolutionary operators are partially based on predicted fitness values. Note that, at the beginning of evolution, the population is evaluated in the real environment to collect necessary state data. Algorithm \ref{alg:gen_based_control} shows the pseudo-code of the generation-based control. 

Previous studies have provided some theoretical analyses of the convergence of surrogate-assisted EAs \cite{chen2013can,queipo2005surrogate}. Here, we demonstrate how the evolution of parameters is affected by different management strategies. As shown in Figure \ref{gen-indi}, inaccuracy may be introduced by the surrogate model under the generation-based control, misguiding the update direction. However, it can be corrected appropriately based on real fitness evaluations, as the SC switches between the real and the predicted fitness functions.

\begin{algorithm}[t]
\caption{Individual-based control}
\label{alg:indi_based_control}
\textbf{Input}: Policy set of population $\mu_{pop}=\{\mu_{1},\mu_{2},...,\mu_{n}\}$; 
\newline \hspace*{0.95cm} The surrogate model $Q_{rl}$; The RL actor $\mu_{rl}$;
\newline \hspace*{0.95cm} Replay buffer $\mathcal{R}$ for RL agent training;
\newline \hspace*{0.95cm} Candidate population size $n^{*}$; Scaling factor $\sigma$.\\
\textbf{Output}: New population $\mu_{pop^{*}}$ 
\begin{algorithmic}[1]
\FOR{$i=1$ to $n^{*}-n$}
\STATE Sample $\epsilon_{i} \sim \mathcal N(0,1)$
\STATE Inject a new candidate ($\mu_{rl}+\sigma \epsilon_{i}$) to $\mu_{pop}$ 
\ENDFOR
\STATE $F_{pre}$ = Surrogate-assisted evaluation($\mu_{pop}$, $Q_{rl}$, $\mathcal{R}$)
\STATE Retain the best $n-1$ actors in $\mu_{pop}$ according to $F_{pre}$ 
\STATE Inject the recorded elite actor to $\mu_{pop}$
\STATE $F_{real}$ = Evaluation($\mu_{pop}$)
\STATE Copy the elite actor
\STATE $\mu_{pop^{*}}$ = Evolutionary methods($F_{real}$, $\mu_{pop}$)
\STATE \textbf{return} New population $\mu_{pop^{*}}$ 
\end{algorithmic}
\label{indi_based}
\end{algorithm}
\subsubsection{Individual-based control}\label{indie}
In our individual-based approach, the evolutionary operators work on real fitness. Assume that the population size is $n$. Before the population is evaluated in the real environment, a candidate population with $n^{*}$ offspring is generated with $n^{*}>n$. After being evaluated by the surrogate model, only the best $n$ individuals are presented to the real environment. This method is also referred to as “preselection strategy”. In principle, to generate the candidate population, apart from the original population, $n^{*}-n$ extra individuals are produced by adding Gaussian noise to the RL actor or the best actor found so far (Algorithm \ref{indi_based}). By default, we mutate the RL actor to better explore its surrounding landscape. 

As shown in Figure \ref{gen-indi}, the surrogate model preselects those mutated individuals with relatively higher predicted fitness and filters out any solution that are likely to fail, then the preselected genetic population actually forms a “trust region” \cite{powell2010convergence}. The real fitness evaluation is finally conducted for stable optimization, resulting in a more directional and smooth parameter update path. 

\subsubsection{Elite protection}
The evolution parts of previous hybrid methods such as ERL and PDERL follow the spirit of elitism mechanism \cite{khadka2018evolution}, where the selected elites with high fitness are kept in the population. In practice, the elites copy a part of their genes to other individuals and are protected from mutations.

Apart from the elitism mechanism within hybrid frameworks, another issue may arise when the surrogate model is used for fitness evaluation. As discussed in Section \ref{critic-surrogate}, inevitably, the ranking of individuals based on the predicted fitness may not perfectly align with that based on the real fitness, resulting in instability of evolution and fluctuation of performance.

To handle this issue, in the generation-based control, SC keeps track of the current best actor when the population is evaluated by the real fitness function, and makes sure it stays in the population after performing evolutionary operators based on predicted fitness. In the individual-based control, the top $n-1$ actors from the candidate population and the current best actor constructs a new population to be applied to the real environment. Our experiment in Section \ref{ep} shows that the elite protection mechanism effectively prevents the dramatic fluctuation of performance and the spread of detrimental information from individuals with inaccurate fitness while using the fixed evolution control. 

\subsection{Evaluation memory}\label{evalmem}
Off-policy DRL methods such as DQN \cite{mnih2013playing}, DDPG \cite{lillicrap2015continuous}, and TD3 \cite{fujimoto2018addressing} maintain a constantly updated replay buffer to improve the sample efficiency of the RL agent. In our approach, the most recent part of the historical data (state information only), referred to as evaluation memory, is exploited to evaluate the population. This memory not only contains diverse state samples but also keeps track of the current optimization process. In addition, evaluation memory does not have to be maintained all the time. It is only created when the surrogate model is called for evaluating individuals and has a negligible memory footprint.

\section{Experiments and Evaluations}\label{experiment}
To highlight the value of SC, we focus on the implementation of SC with two state-of-the-art hybrid frameworks ERL and PDERL, referred to as SERL and SPDERL, respectively. We aim to answer the following questions: (1) Does SC improve the computational efficiency and the performance of the original hybrid frameworks? (2) How sensitive is the optimization process to the control parameters of SC's management strategies? (3) What impacts does SC bring to the internal dynamics of the original hybrid frameworks?
\begin{table}[!h] 
\centering 
\caption{Action and state dimensions in various environments} 
\label{tab:state-action-env}
\fontsize{9}{9}\selectfont    
\begin{threeparttable} 
\begin{tabular}{c|cc}  
\toprule         
{\bf Environment}&{\bf State dimension}&{\bf Action dimension}\cr
\midrule 
Ant & $111$ & $8$ \cr
Hopper & $11$ & $3$ \cr
Walker & $17$ & $6$ \cr
Swimmer & $8$ & $2$  \cr
Reacher & $11$ & $2$ \cr
HalfCheetah & $17$ & $6$ \cr
\bottomrule 
\end{tabular}
\end{threeparttable}
\end{table}

\subsection{Environmental settings}
We performed experiments on 6 continuous control tasks: HalfCheetah, Ant, Hopper, Swimmer, Reacher and Walker with the MuJoCo\footnote{https://mujoco.org/} physics engine \cite{todorov2012mujoco}, and Table \ref{tab:state-action-env} shows the action and state dimensions in all environments. All these tasks are packaged according to the standard OpenAI Gym API\footnote{https://gym.openai.com/} and friendly to simulation. At the beginning of each simulation, an initial state vector determined by internal random seeds is provided, and at each subsequent time step, the policy neural network calculates what action to perform according to the latest state vector. The environment simulates this action and returns a new state vector and the corresponding reward. Additionally, the reward function is task-specific and we consider the real fitness value as the sum of the rewards over one episode or the average reward over several runs.

\begin{table*} 
\centering 
\caption{Summary of the baselines and the proposed algorithms} 
\label{tab:algo-settings}
\fontsize{9}{9}\selectfont   
\begin{threeparttable} 
\begin{tabular}{c|cccc}  
\toprule        

{\bf Algorithms}&{\bf EA Part}&{\bf RL Part}&{\bf Surrogate Fitness}&{\bf Management Strategies}\cr
\midrule 
ERL & GA (N-point Crossover \& Gaussian Mutation) & DDPG & No & No\cr
SERL-I & GA (N-point Crossover \& Gaussian Mutation) & DDPG & Yes & Individual-based Control\cr
SERL-G & GA (N-point Crossover \& Gaussian Mutation) & DDPG & Yes & Generation-based Control\cr
PDERL & GA (Distillation Crossover \& Proximal Mutation) & DDPG & No & No\cr
SPDERL-I & GA (Distillation Crossover \& Proximal Mutation) & DDPG  & Yes & Individual-based Control\cr
SPDERL-G & GA (Distillation Crossover \& Proximal Mutation) & DDPG  & Yes & Generation-based Control\cr
\bottomrule 
\end{tabular}
\end{threeparttable}
\end{table*}

\begin{table} 
\centering 
\caption{Hyperparameters of SERL and SPDERL} 
\label{tab:Hyperparameters of ERL and PDERL}
\fontsize{9}{9}\selectfont    
\begin{threeparttable} 
\begin{tabular}{lc}  
\toprule         
{\bf Hyperparameter}&{\bf Value}\cr
\midrule
Hidden layers of the actor network & $(64,64)$\cr
Hidden layers of the critic network & $(400,300)$\cr
Activation function of the actor network & Tanh\cr
Activation function of the critic network & ELU\cr
Target weight $\tau$ & $0.001$\cr
RL actor learning rate & $5e^{-5}$ \cr
RL critic learning rate & $5e^{-4}$ \cr
Replay buffer size & $1e^{6}$\cr
RL agent batch size  & $128$ \cr
Discount factor & $0.99$ \cr
Optimizer & Adam \cr
Genetic actor learning rate & $1e^{-3}$ \cr
Genetic memory size & $8000$ \cr
Population size & $10$ \cr
Genetic agent crossover batch size & $128$\cr
Genetic agent mutation batch size & $256$ \cr
Distillation crossover epochs  & $12$ \cr
Mutation probability & $0.9$ \cr
Mutation strength & $0.1$ \cr
\bottomrule 
\end{tabular}
\end{threeparttable}
\end{table}

\begin{table}[h]
\centering 
\caption{Hyperparameters used in various environments} 
\label{tab:Hyperparameters env}
\fontsize{9}{9}\selectfont    
\begin{threeparttable} 
\begin{tabular}{c|lccc}  
\toprule         
{\bf Environment}&{\bf Algorithm}&{\bf Elite }&{\bf Trials}&{\bf Sync}\cr
\midrule 
\multirow{2}{*}{Ant} & SERL & $0.3$ & $1$ & $1$\cr
& SPDERL & $0.2$ & $1$ & $1$\cr
\midrule 
\multirow{2}{*}{Hopper} & SERL & $0.3$ & $5$ & $1$\cr
& SPDERL & $0.2$ & $3$ & $1$\cr
\midrule 
\multirow{2}{*}{Walker} & SERL & $0.2$ & $3$ & $1$\cr
& SPDERL & $0.2$ & $5$ & $1$\cr
\midrule 
\multirow{2}{*}{Swimmer} & SERL & $0.1$ & $1$ & $10$\cr
& SPDERL & $0.1$ & $1$ & $10$\cr
\midrule 
\multirow{2}{*}{Reacher} & SERL & $0.1$ & $1$& $10$\cr
& SPDERL & $0.1$ & $1$ & $10$\cr
\midrule 
\multirow{2}{*}{HalfCheetah} & SERL & $0.1$ & $1$& $1$\cr
& SPDERL & $0.1$ & $1$ & $10$\cr
\bottomrule 
\end{tabular}
\end{threeparttable}
\label{para_various}
\end{table}

\subsection{Algorithm settings}

We use the official implementations of ERL\footnote{https://github.com/ShawK91/Evolutionary-Reinforcement-Learning} \cite{khadka2018evolution} and PDERL\footnote{https://github.com/crisbodnar/pderl} \cite{bodnar2020proximal} as the major baselines and follow all their hyperparameter settings for EAs, RL agents, neural networks and the population size ($n=10$). The actor (agent’s policy) is represented by a fully connected neural network with two hidden layers, each containing 64 neurons. The number of neurons in the input and output layers is task-specific (Table \ref{tab:state-action-env}), and the hidden and output layers take the Tanh activation function. Thus, the policy space is encoded by 5702 parameters in the Walker task, 5702 in HalfCheetah, 5058 in Reacher, 11848 in Ant and 5123 in Hopper. Furthermore, the critic network is also fully connected, and the numbers of neurons in the first and second hidden layers are 400 and 300, respectively. We use the ELU activation between hidden layers. 

The main difference between ERL and PDERL is in the EA part. In ERL, the GA employs the typical N-point crossover and Gaussian mutation, while in PDERL, the crossover is implemented by distillation, and the Proximal mutation, based on the SM-G-SUM operator \cite{lehman2018safe} is used. Table \ref{tab:algo-settings} provides a brief summary of ERL, PDERL and our methods. Moreover, we use a standard Genetic Algorithm \cite{khadka2018evolution} and the DDPG implemented by OpenAI Spinningup \footnote{https://github.com/openai/spinningup} as extra baselines to compare our methods against pure EA and RL algorithms. Table \ref{tab:Hyperparameters of ERL and PDERL} shows the hyperparameters of SERL and SPDERL in this work, they are common over all environments. ``Elite” in Table \ref{para_various} represents the proportion of elite individuals in the genetic population. ``Trials” is referred to as the number of evaluation times of an individual, for Walker and Hopper, where the reward variance is relatively higher than other environments, policies need run more times to obtain their average fitness. Finally, ``Sync” is the synchronization period of the RL-Actor.

For SC's default hyperparameters, the maximum evaluation memory size $k$ is limited to $50,000$, and we fix $\omega$ to 0.6 for the generation-based control, which means that, in each generation, the population has a $60\%$ chance of being evaluated by the surrogate model. For the individual-based control, $\alpha=(n^{*}-n)/n$ is the control factor for the candidate population size with $\alpha=1$ in our experiments and we set the scaling factor $\sigma=0.01$ for generating Gaussian noise. During the training, SC keeps recording the current best actor when the population is evaluated in the real environment for elite protection. The periodical synchronization of the RL actor and the population is performed only after real fitness evaluation. We refer to SERL and SPDERL with generation-based control as SERL-G and SPDERL-G. Similarly, SERL-I and SPDERL-I indicate the combination of SERL and PDERL with individual-based control, respectively. 

\begin{figure*}[t]
\centering
\includegraphics[width=0.33\textwidth]{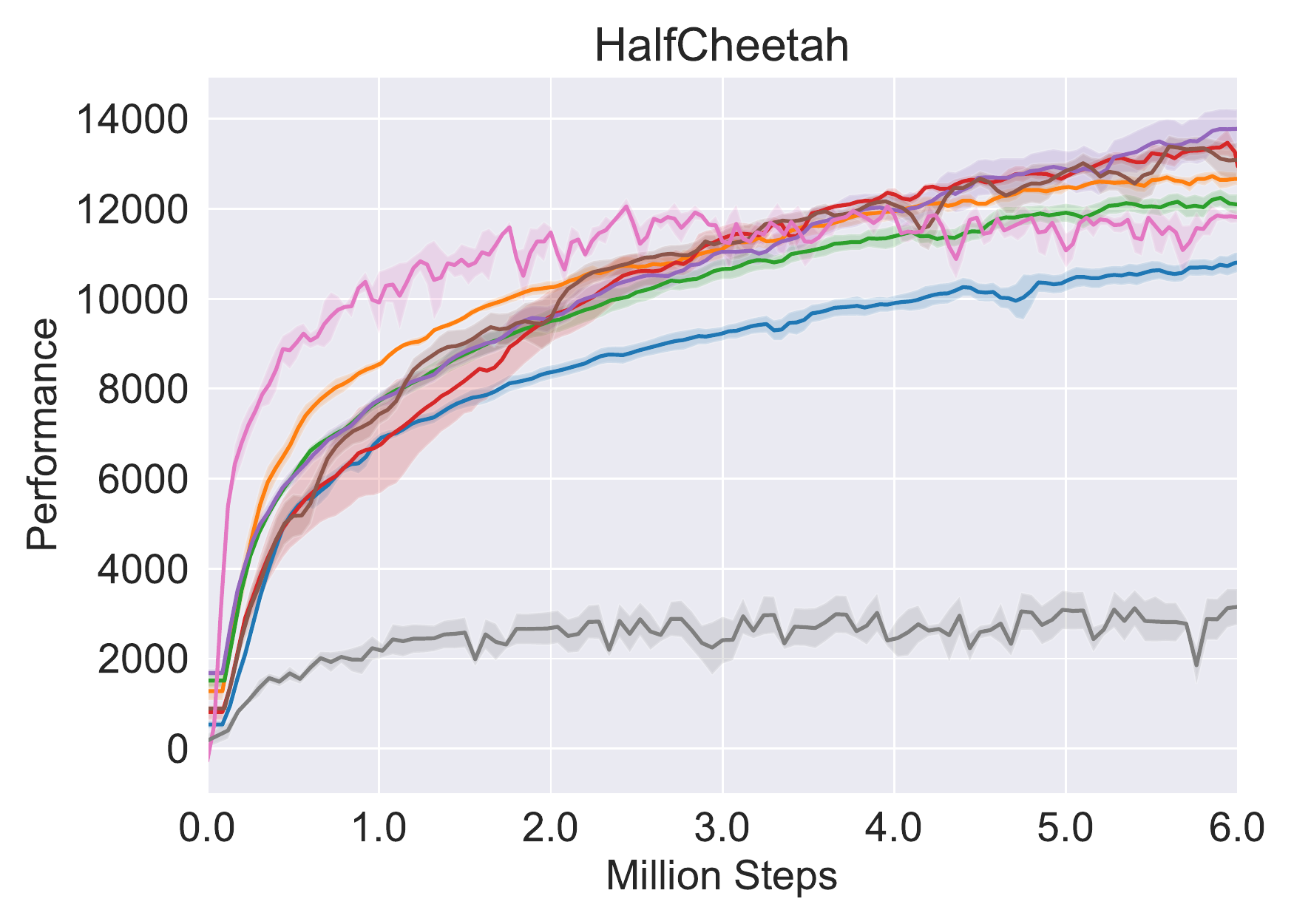}
\includegraphics[width=0.33\textwidth]{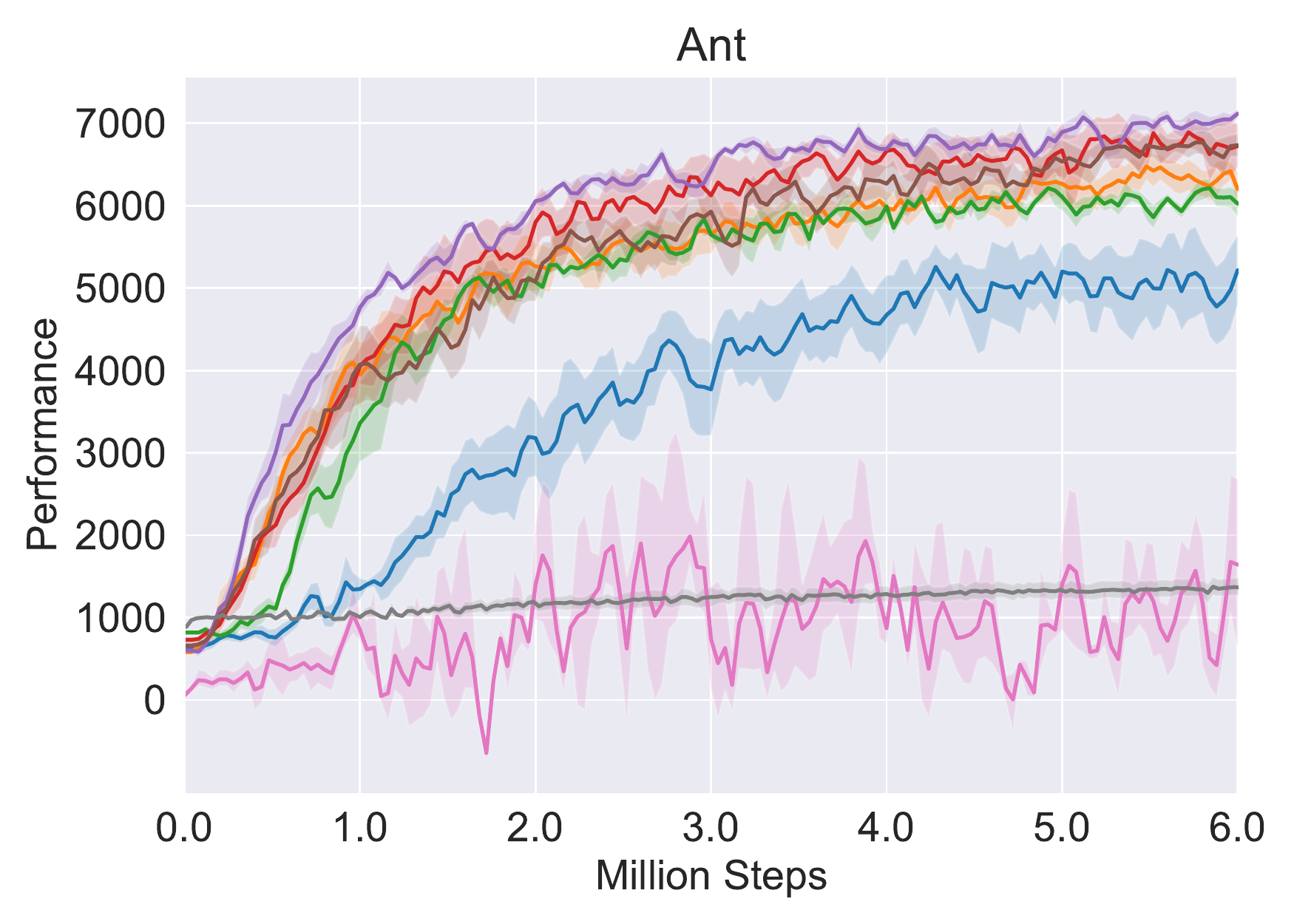} 
\includegraphics[width=0.33\textwidth]{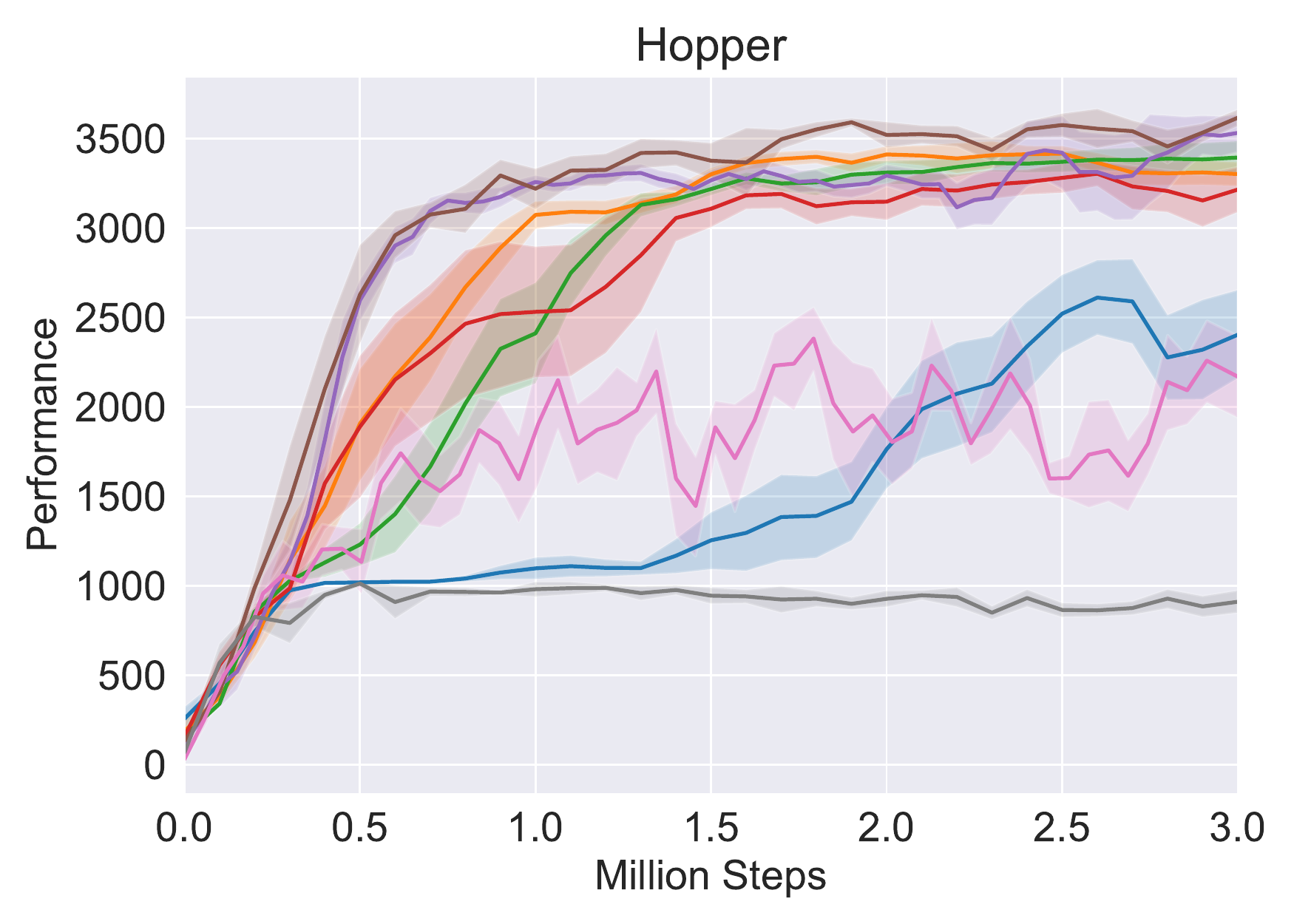} 
\includegraphics[width=0.33\textwidth]{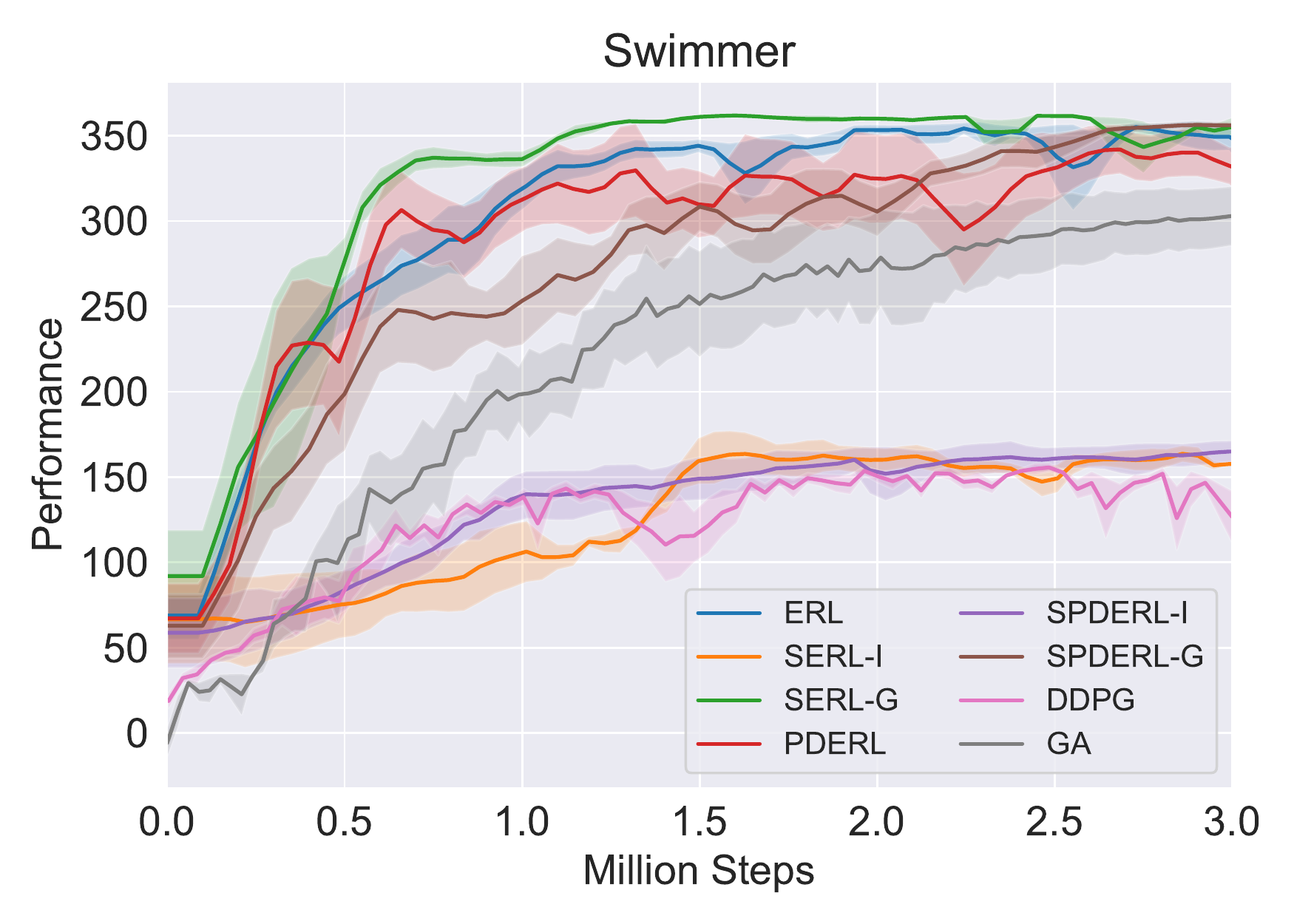}
\includegraphics[width=0.33\textwidth]{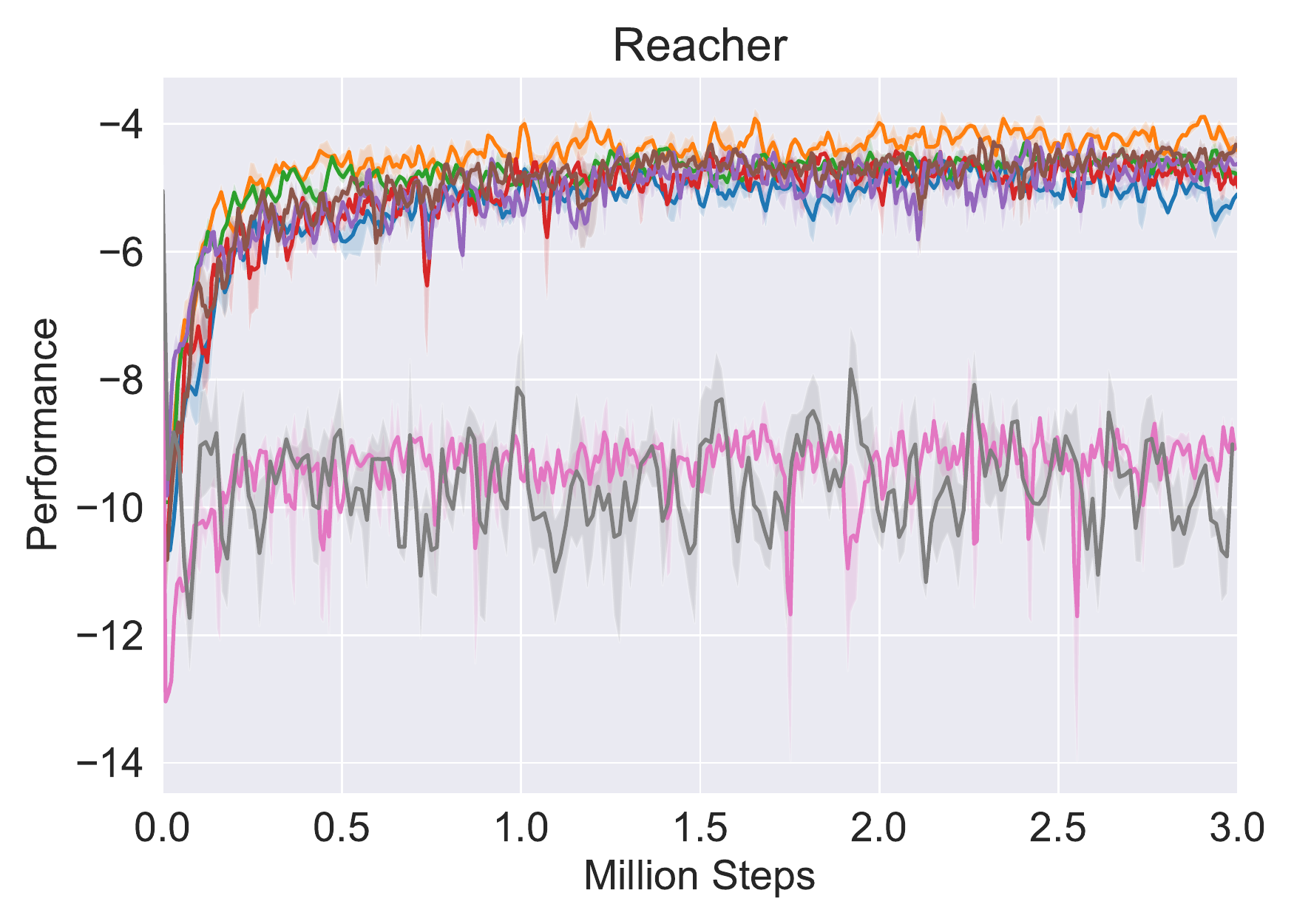} 
\includegraphics[width=0.33\textwidth]{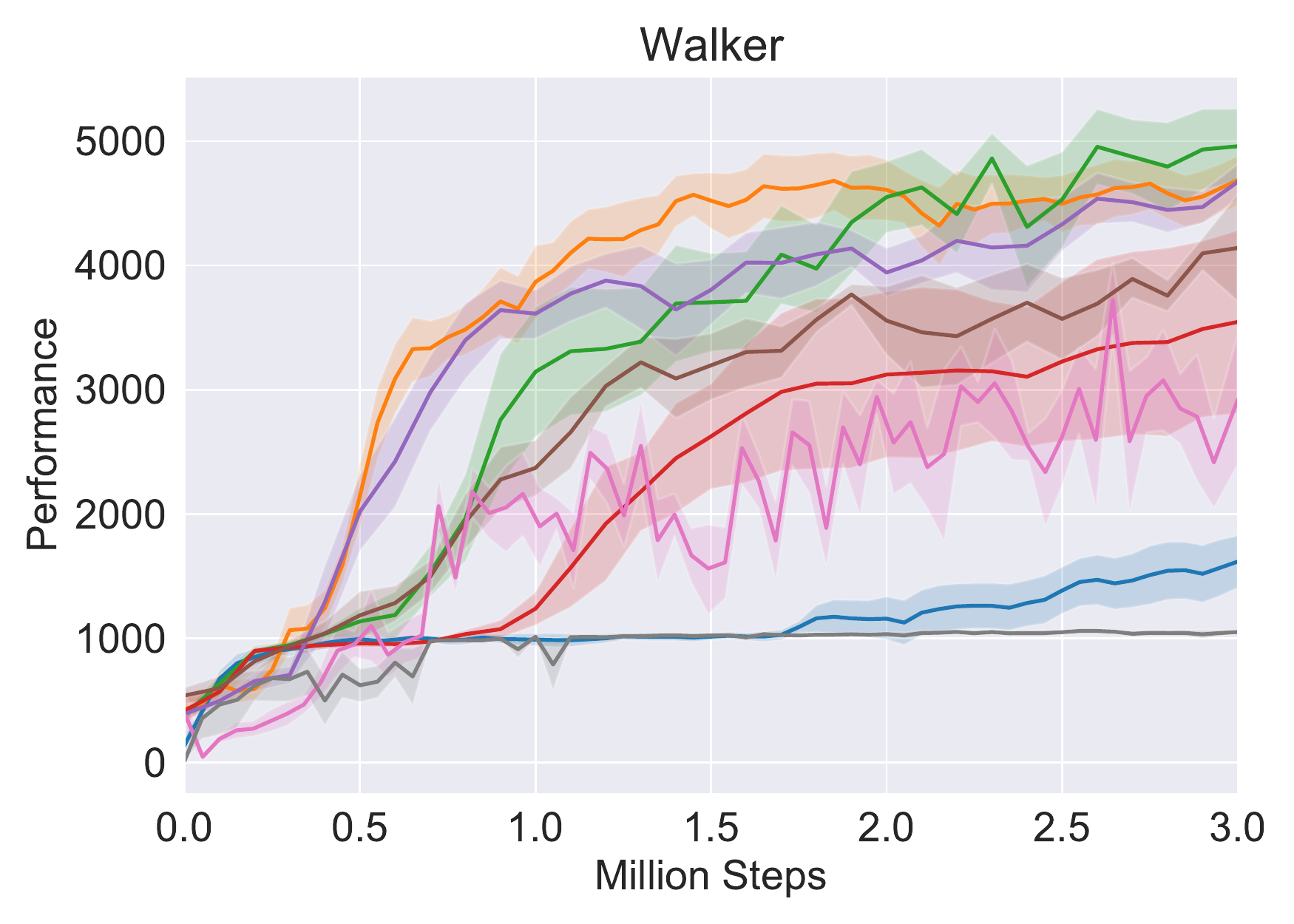} 
\caption{Learning curves on six MuJoCo environments: HalfCheetah, Ant, Hopper, Swimmer, Reacher and Walker.}
\label{fitness}
\end{figure*}

\subsection{Overall performance}\label{overall}
We train each proposed method with 6 different random seeds on 6 MuJoCo environments following the convention in literature \cite{todorov2012mujoco}. During the training process, the average of 5 test results of the best actor from the genetic population is reported as the performance of each algorithm. Figure \ref{fitness} illustrates their learning processes during training. The solid curves represent the mean values and the shaded regions indicate the standard deviations. 

In most environments, SC can significantly improve the learning speed and the performance of the original hybrid frameworks, and also make the learning process more stable with lower variance. For example, SERL-G can outperform ERL across all environments. Except for Swimmer, the improvement of SERL-I is more evident in the early training phase (within 1 million steps) compared to SERL-G, and it outperforms other methods in Reacher. When it comes to SPDERL, both SPDERL-I and SPDERL-G outperform PDERL in Hopper and Walker. For Ant and HalfCheetah, SPDERL-I can achieve higher final performance while SPDERL-G can accelerate the performance improvement in the early training phase. As for pure RL and EA methods, GA struggles in most environments except Swimmer. In HalfCheetah, DDPG performs better than any other methods in the early training phase but the performance tends to converge after 3 million steps.  

It is worth noting that DDPG and all methods under the individual-based control fail in Swimmer and, as explained in \cite{khadka2018evolution,pourchot2018cem}, DRL methods face great challenges in effectively learning the gradient information. To alleviate this issue, we generate the candidate population by mutating the best actor that has been found by genetic operations (Appendix A). In this specific environment, the evolutionary search is more suitable for driving the optimization process.

\subsection{Parameter analysis}
\begin{figure*}[t]
\centering
\includegraphics[width=0.245\textwidth]{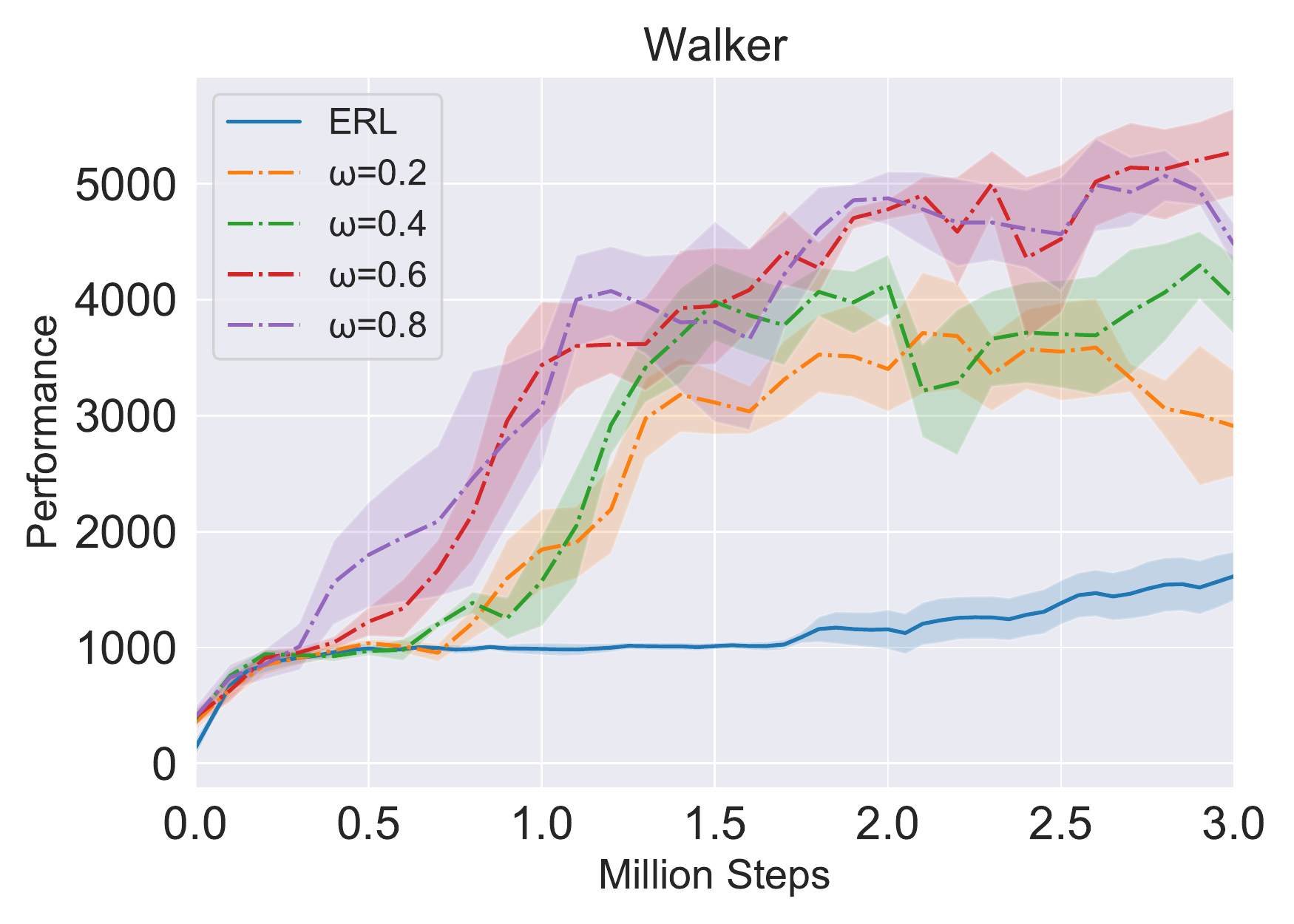} 
\includegraphics[width=0.245\textwidth]{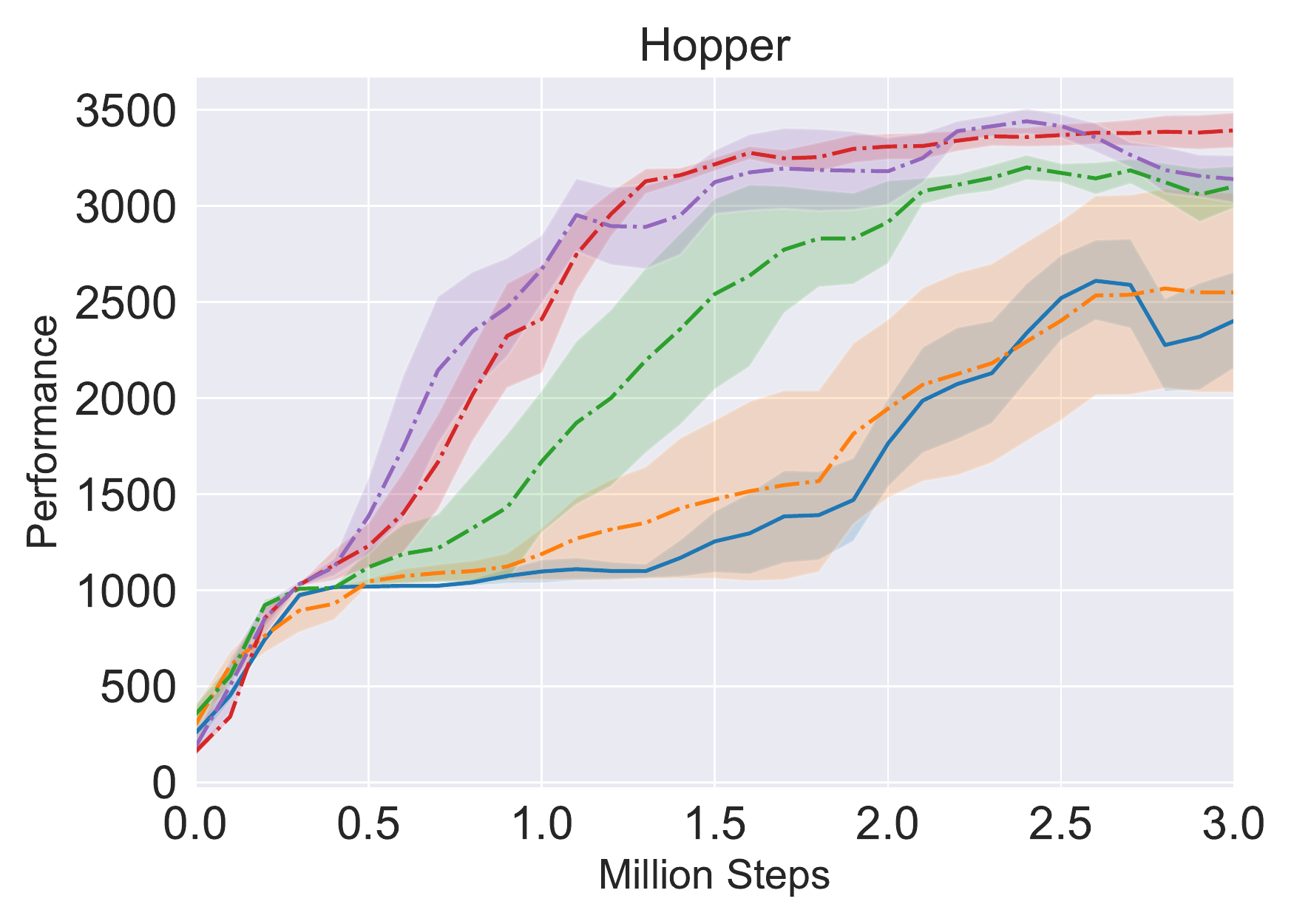}
\includegraphics[width=0.245\textwidth]{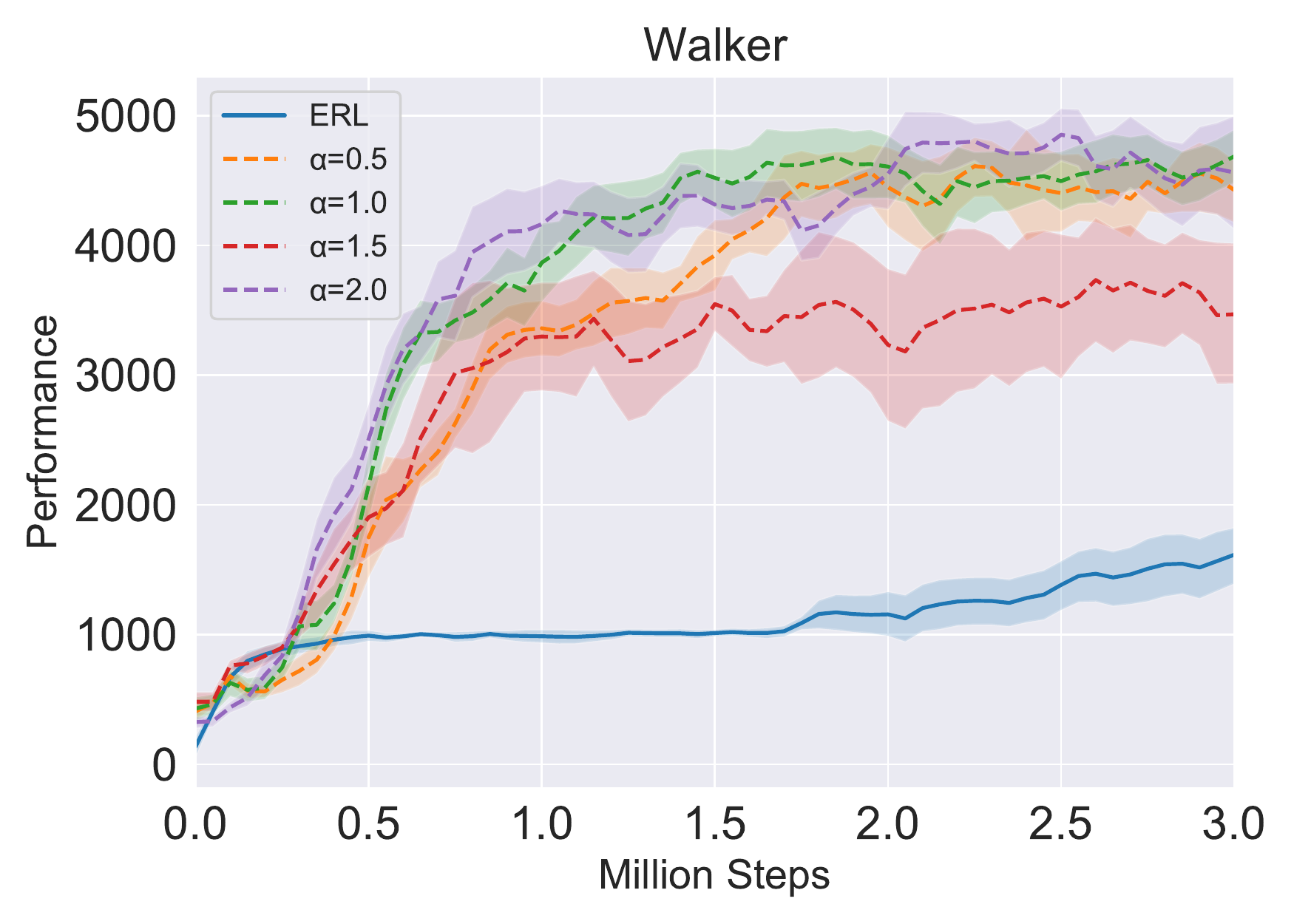} 
\includegraphics[width=0.245\textwidth]{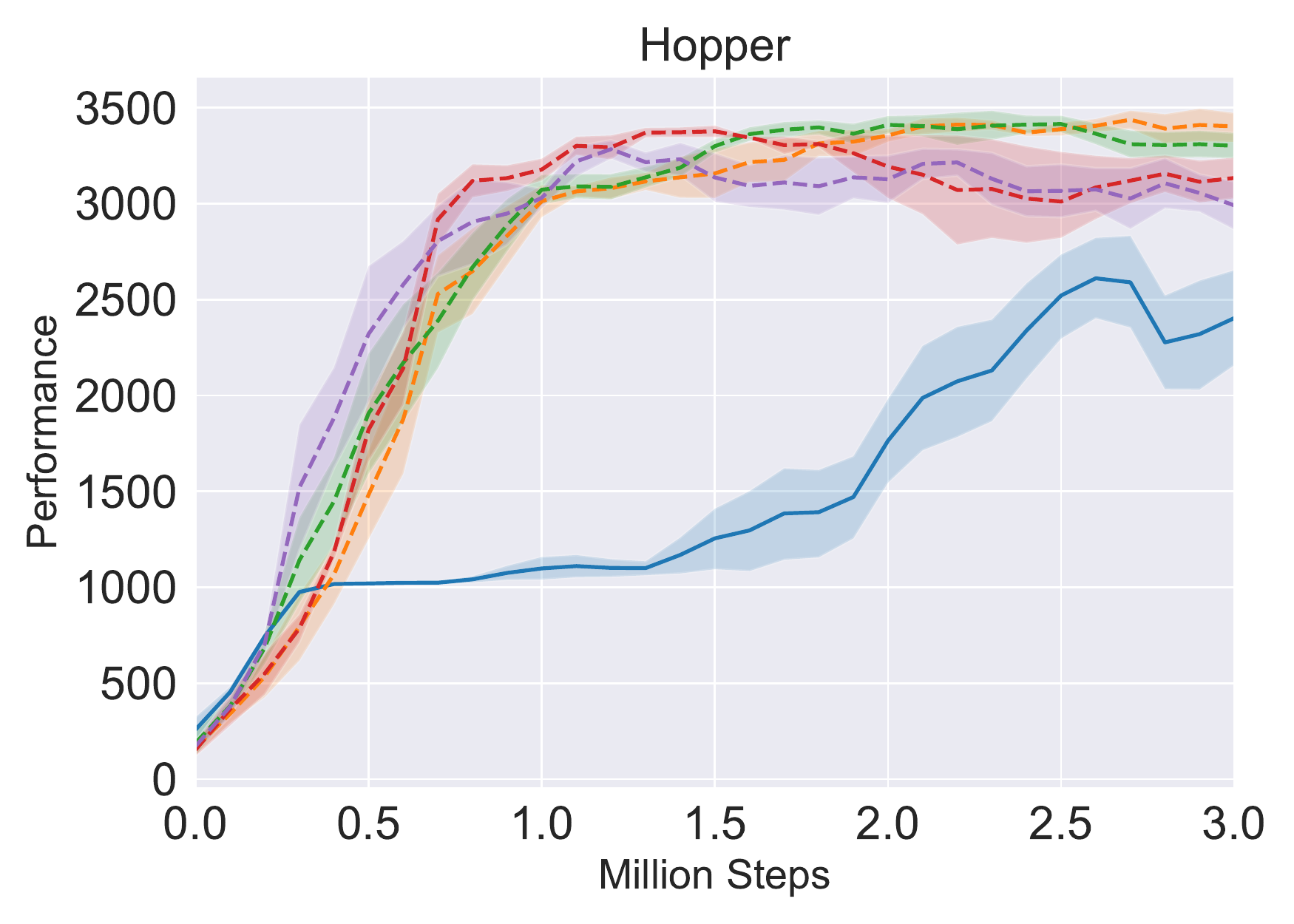}
\includegraphics[width=0.245\textwidth]{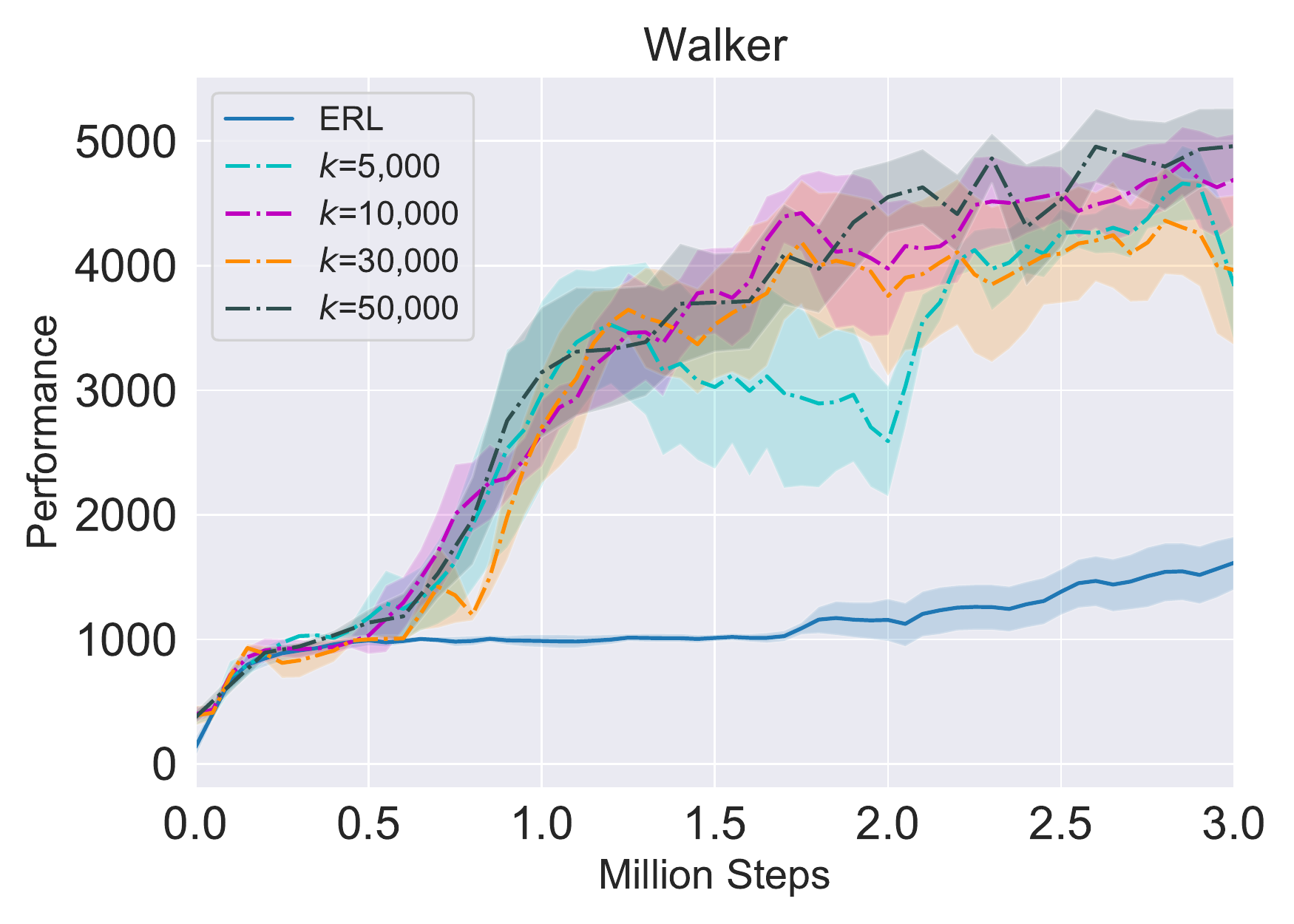} 
\includegraphics[width=0.245\textwidth]{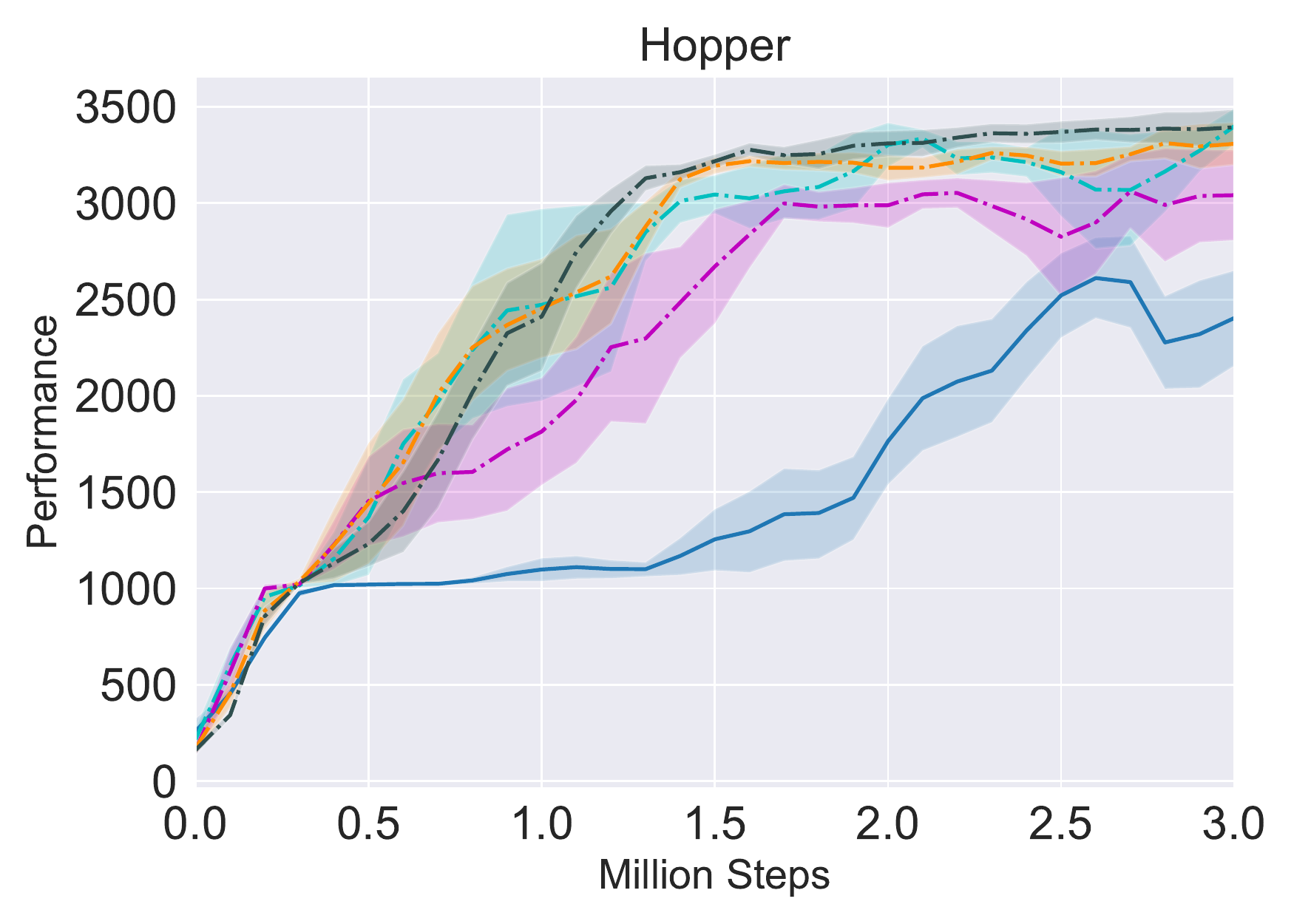}
\includegraphics[width=0.245\textwidth]{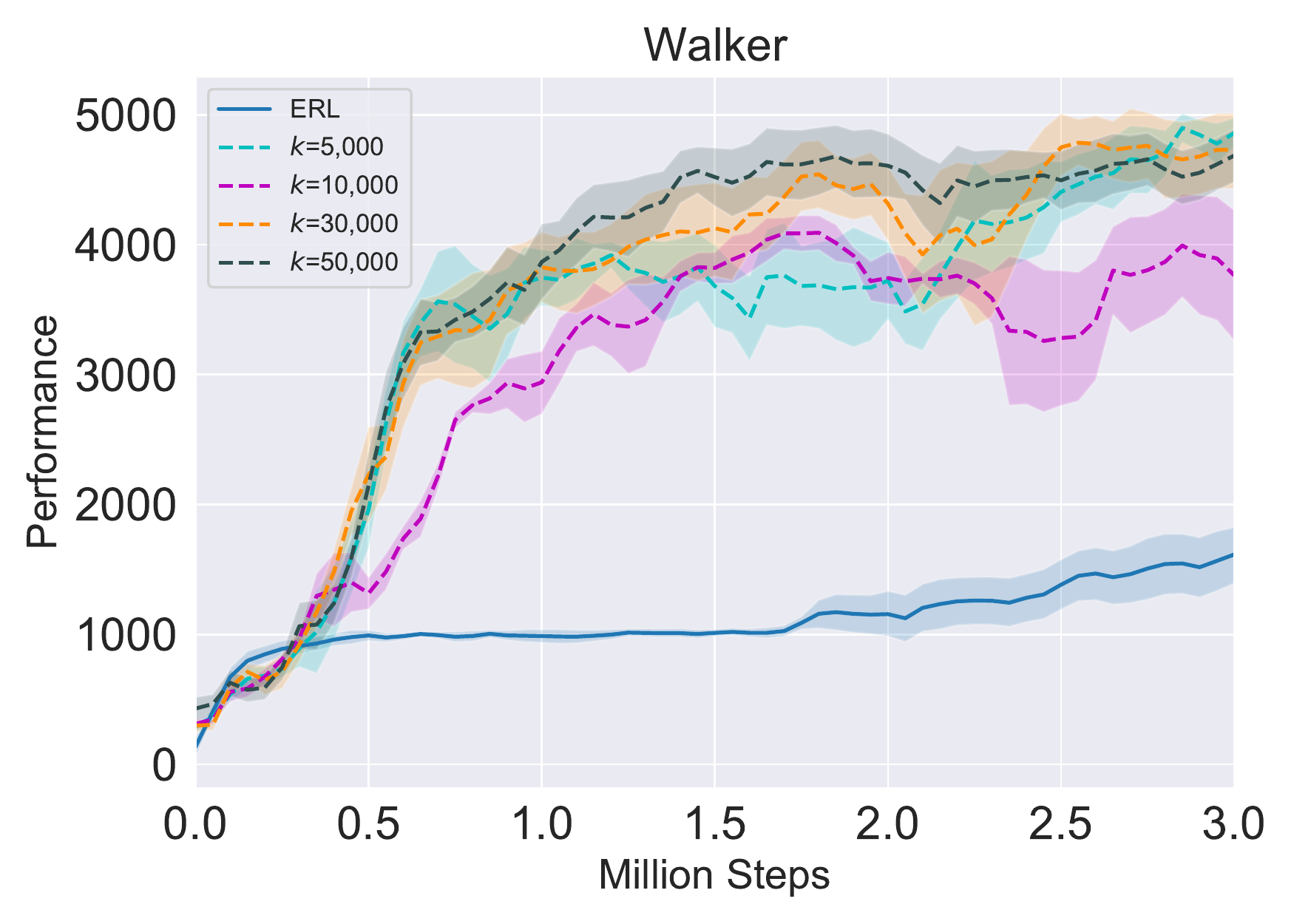}
\includegraphics[width=0.245\textwidth]{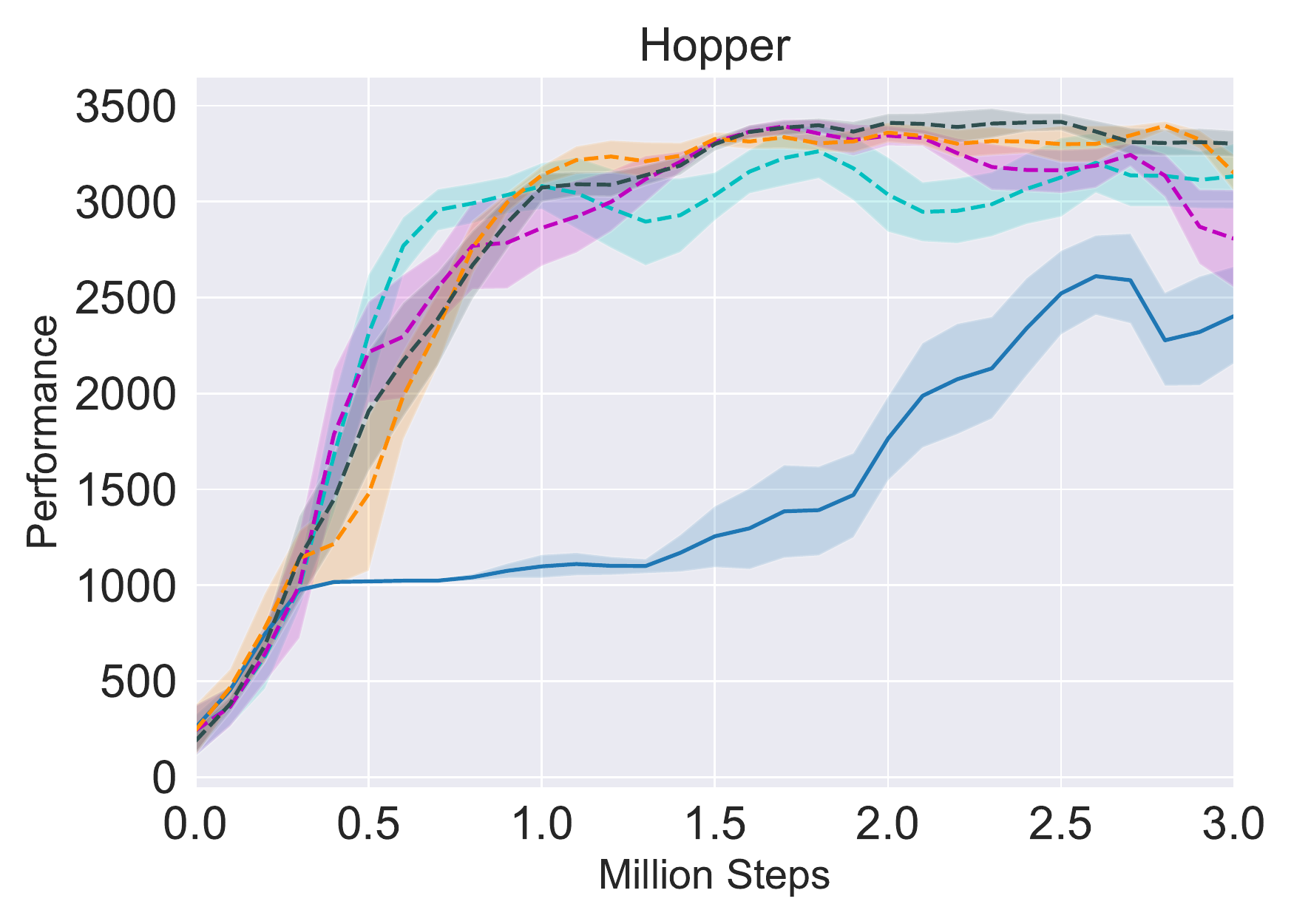}
\caption{Parameter analysis of SERL-G with different generation-based control factors and evaluation memory sizes (first two columns) and SERL-I with different individual-based control factors and evaluation memory sizes (last two columns) in Walker and Hopper environments.}
\label{fig5}
\end{figure*}
In this part, we investigate the influence of the following parameters: the ratio of surrogate-assisted evaluation $\omega$, the control factor $\alpha$ of the candidate population size, and the capacity $k$ of evaluation memory, as shown in Figure \ref{fig5}.

\textbf{Control factors.} We vary $\omega$ from $0.2$ to $0.8$ in SERL-G and $\alpha$ from $0.5$ to $2.0$ in SERL-I, respectively. The results indicate that both the final performance and the learning speed are generally improved by increasing $\omega$ under the generation-based control. Although a high value (e.g., $\omega=0.8$) may lead to a little drop in the final performance, it significantly reduces the number of interactions with the environment and speeds up the learning process. For individual-based control, a relatively small $\alpha$ value (e.g., $\alpha=0.5$) is more cost-effective. As the surrogate model needs to evaluate additional individuals generated by mutations, a high value of $\alpha$ may result in overhead, especially when the input of the surrogate model is high-dimensional.

\textbf{Memory capacity.} Furthermore, we investigate the impact of the capacity of the evaluation memory. The second row of Figure \ref{fig5} shows the performance of SERL-G and SERL-I with fixed control factors $\omega=0.6$, $\alpha=1.0$ and various $k$ values in two environments. Overall, a large evaluation memory contains more diverse state data and significantly helps improve the quality of the evaluation, but it also increases the computational cost of SC. In general, the generation-based control is better suited with large evaluation memories to counteract the bias in surrogate-assisted evaluation. By contrast, since the evolutionary methods are based on real fitness evaluation with low deviations, a relatively small $k$ is suitable for individual-based control.

\subsection{Elite protection evaluation}\label{ep}
\begin{figure*}[t]
\centering
\includegraphics[width=0.245\textwidth]{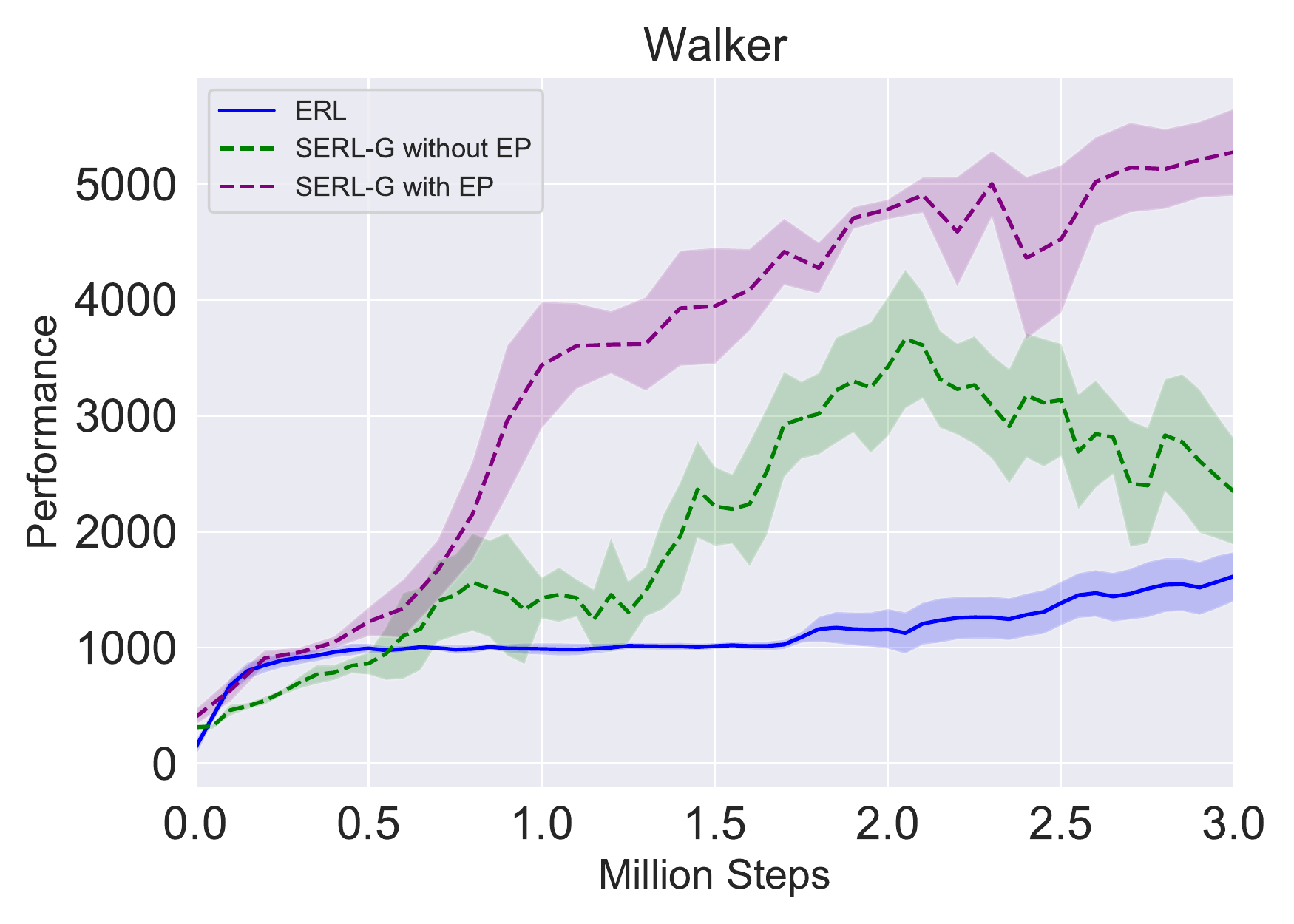} 
\includegraphics[width=0.245\textwidth]{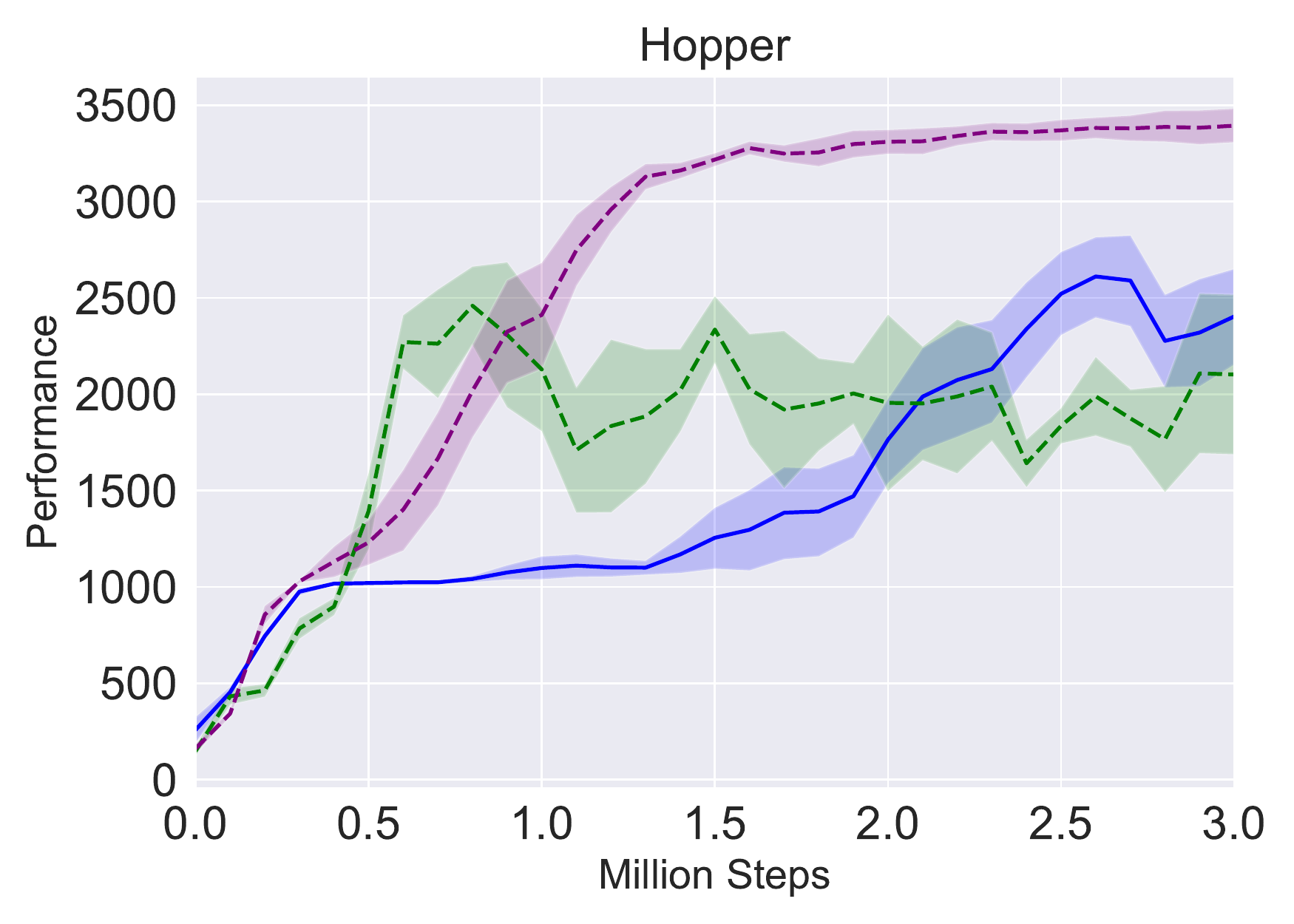}
\includegraphics[width=0.245\textwidth]{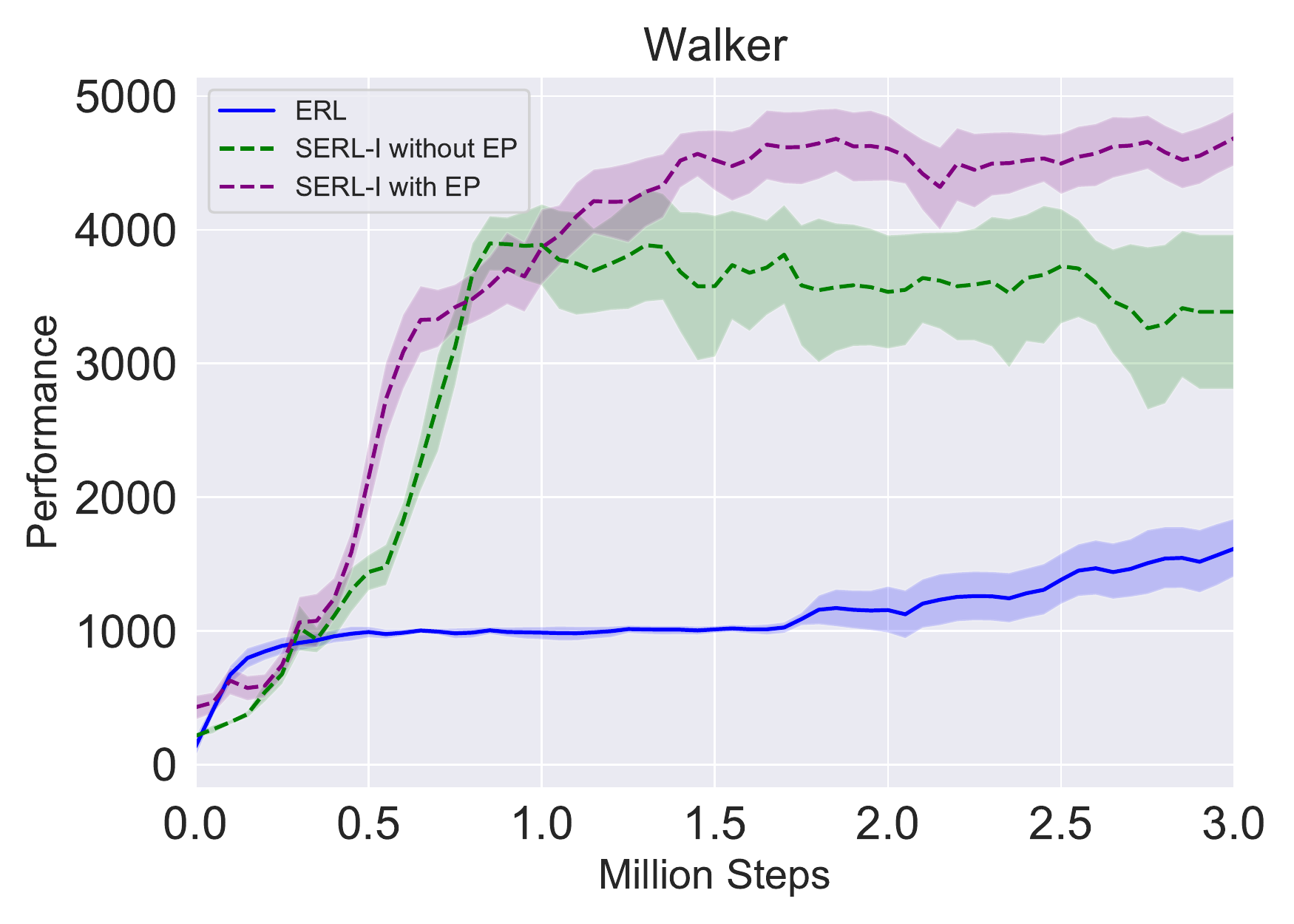} 
\includegraphics[width=0.245\textwidth]{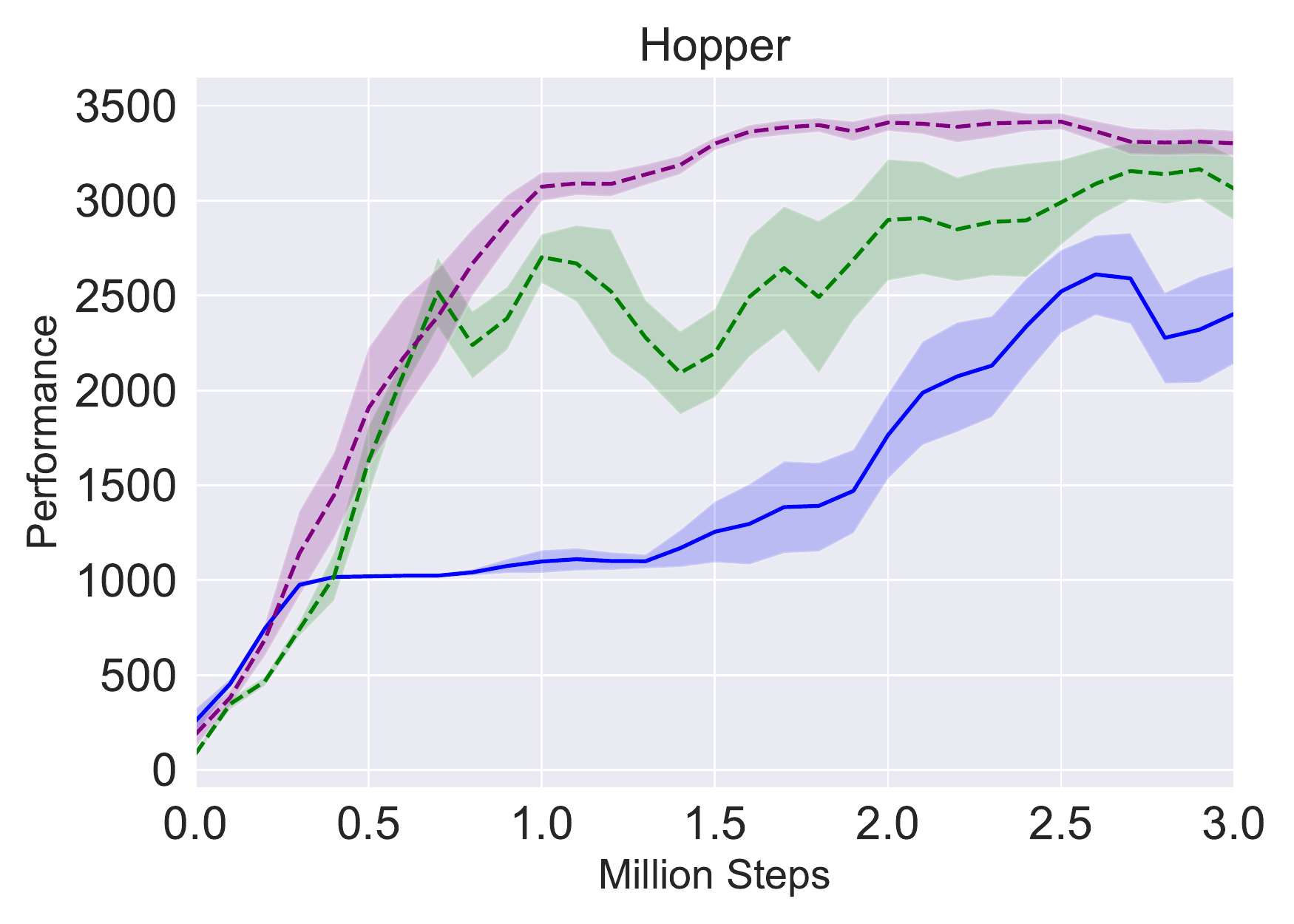}
\caption{Comparisons of SERL-G (first two figures) and SERL-I (last two figures) with and without Elite Protection (EP) mechanism in Walker and Hopper environments.}
\label{fig6}
\end{figure*}
Figure \ref{fig6} shows the performance of SERL-I and SERL-G without the elite protection mechanism. In this setting, SC only speeds up the policy improvement in the early period of the training process and then encounters a dramatic drop in performance. Although the surrogate model can provide a roughly accurate estimation of population fitness, its estimation of elites is possibly biased, which increases the risk of discarding elites from the population. This experiment underlines the importance of elite protection while using the surrogate for fitness evaluation.

\subsection{Changes of the internal dynamics}

\begin{figure*}[t]
\centering

\includegraphics[width=0.33\textwidth]{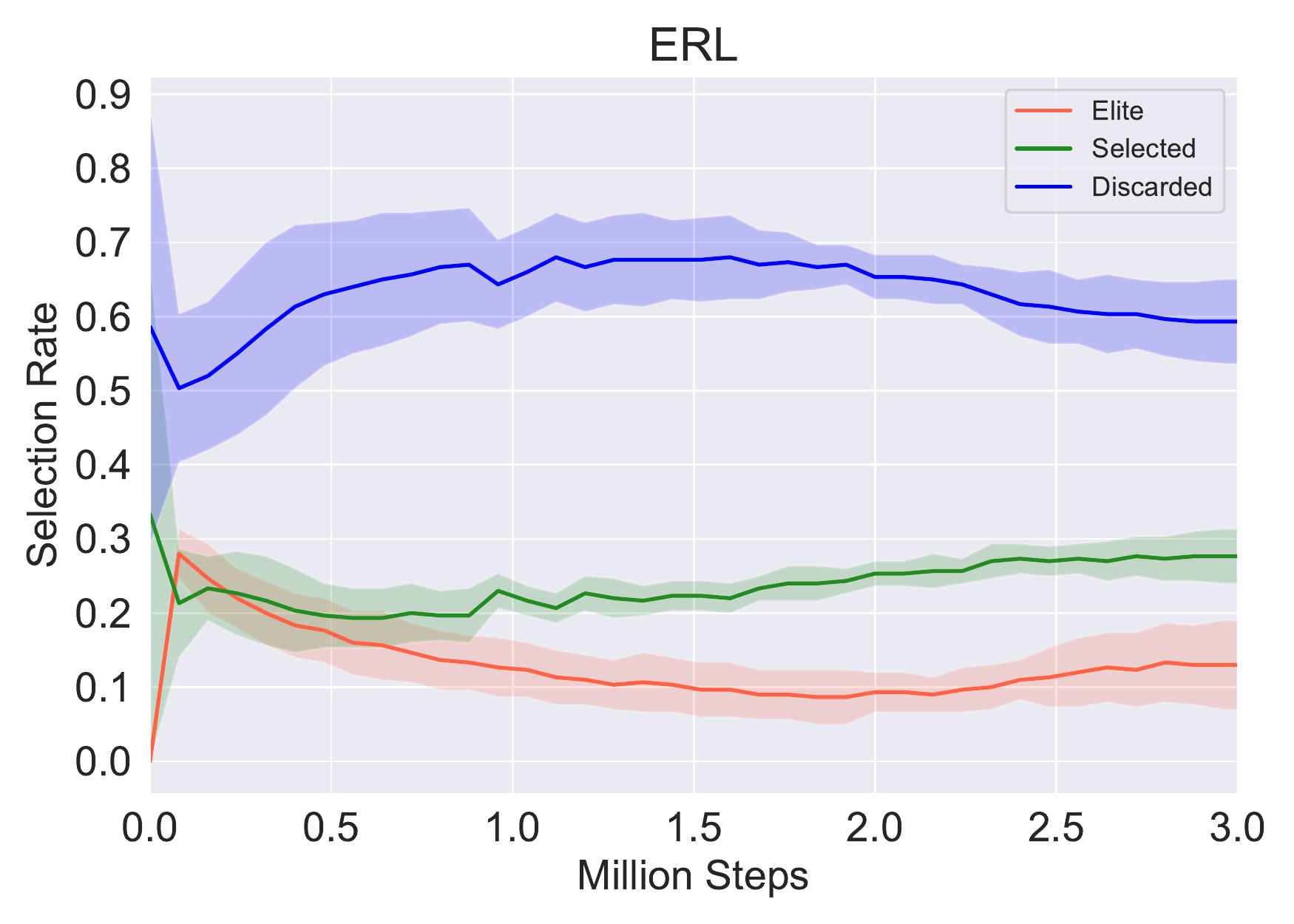}
\includegraphics[width=0.33\textwidth]{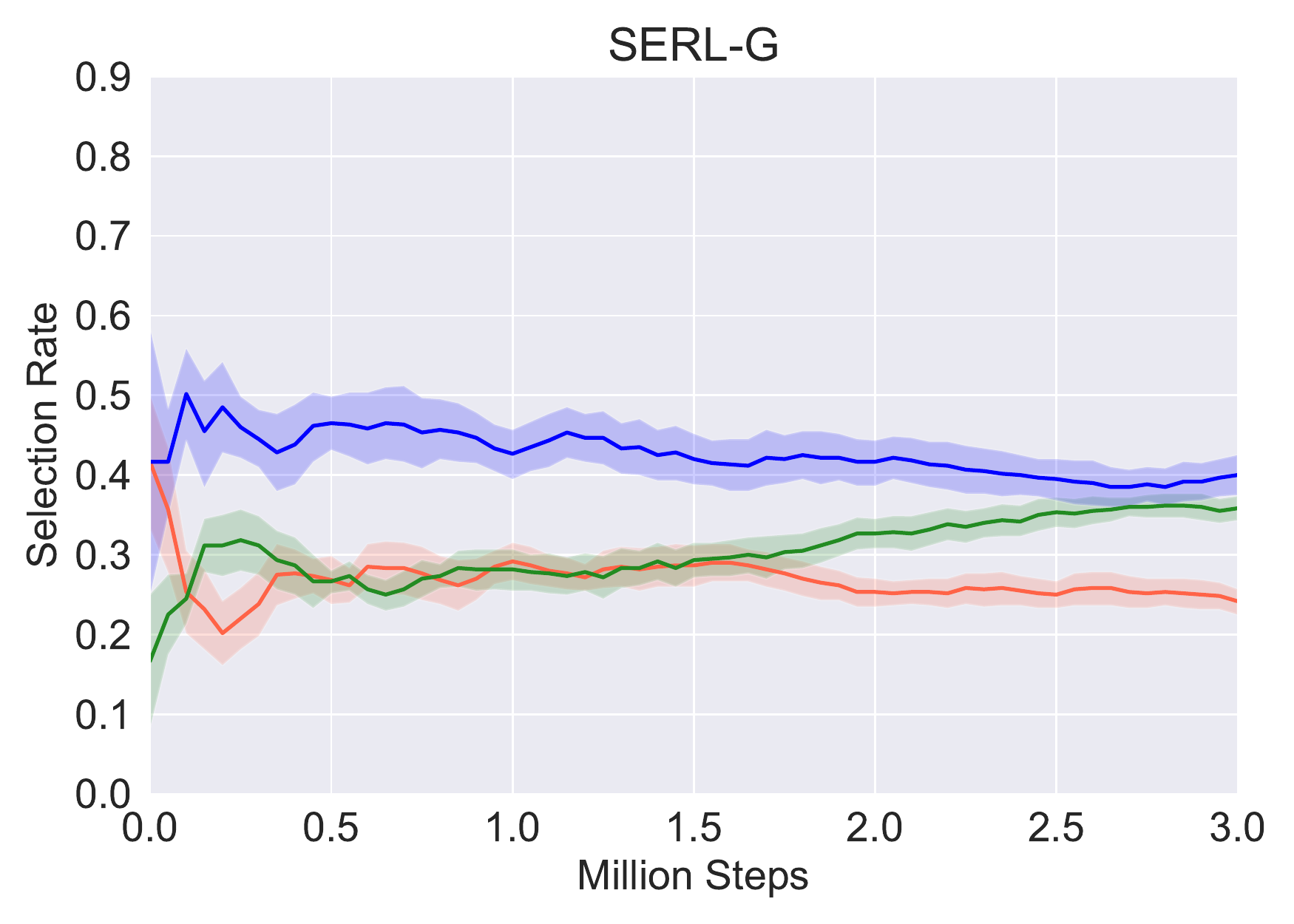} 
\includegraphics[width=0.33\textwidth]{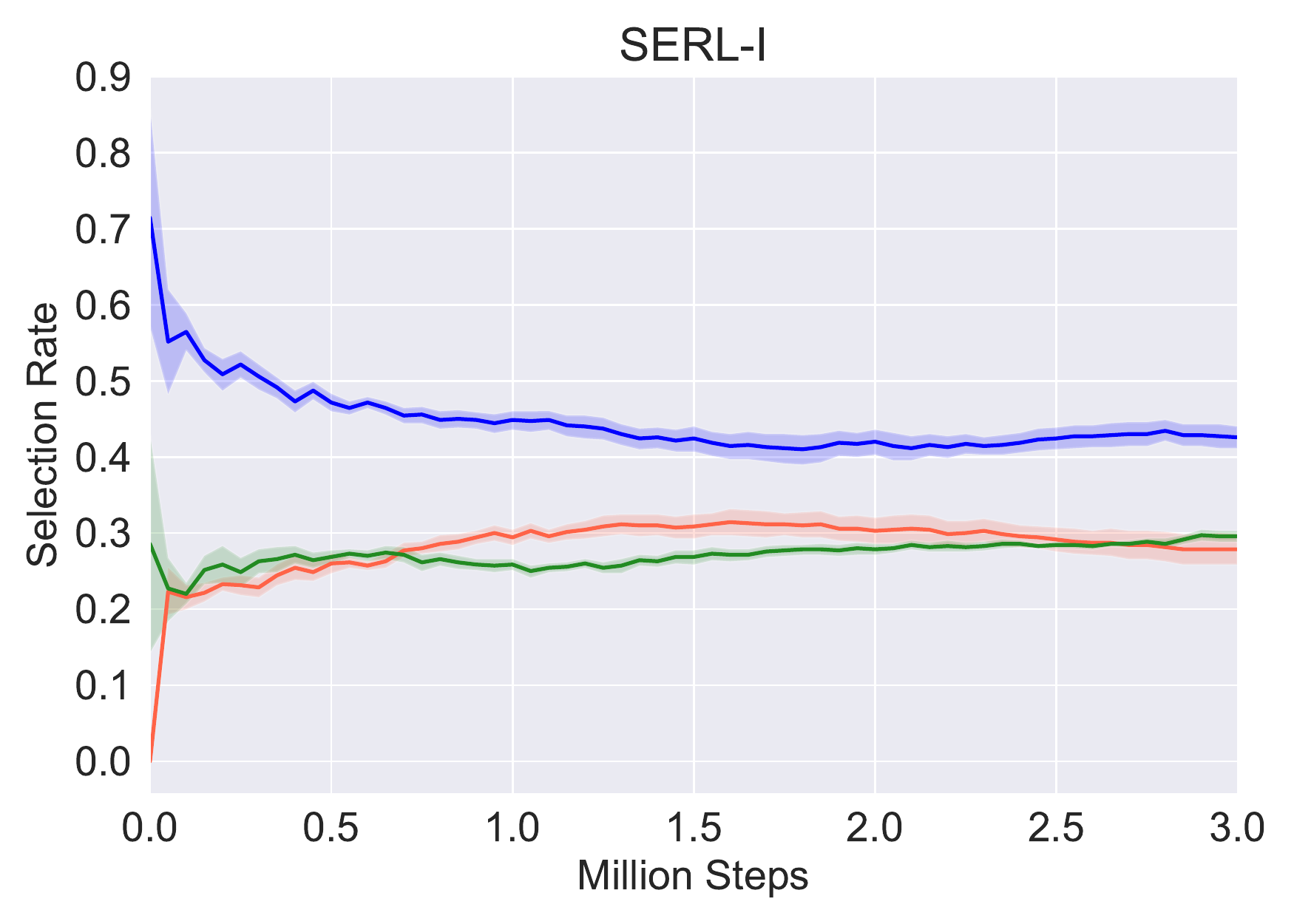} 
\includegraphics[width=0.33\textwidth]{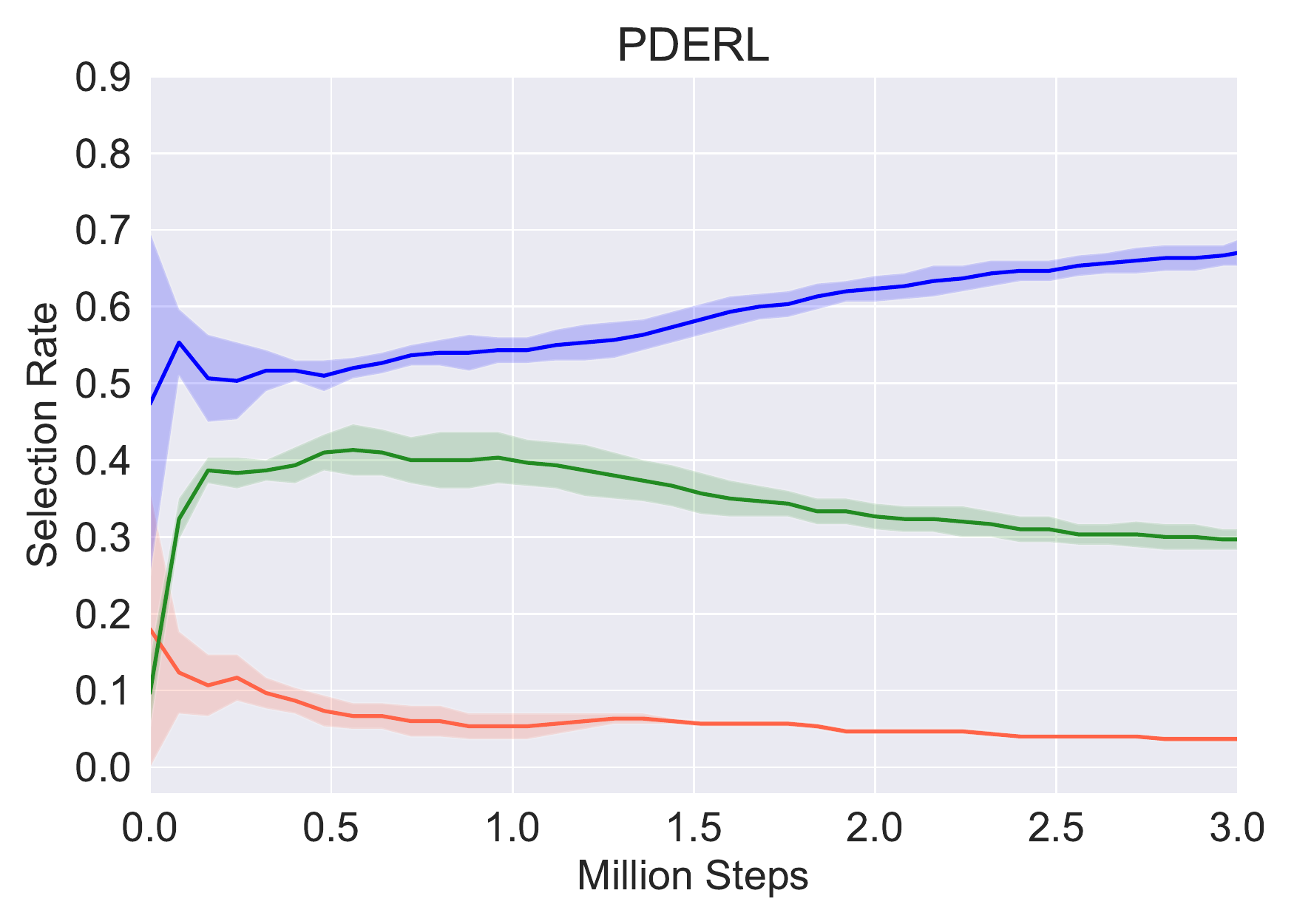}
\includegraphics[width=0.33\textwidth]{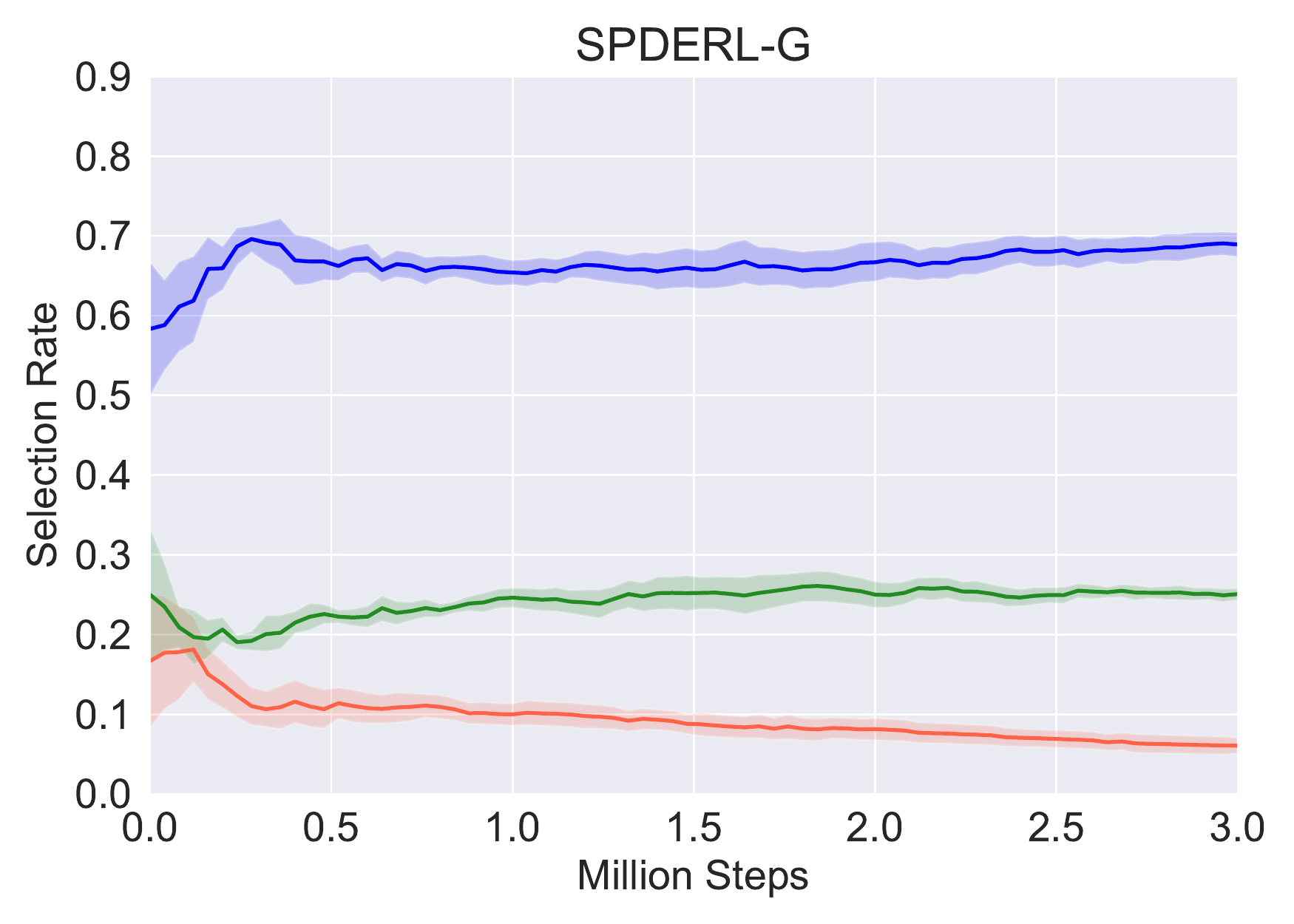} 
\includegraphics[width=0.33\textwidth]{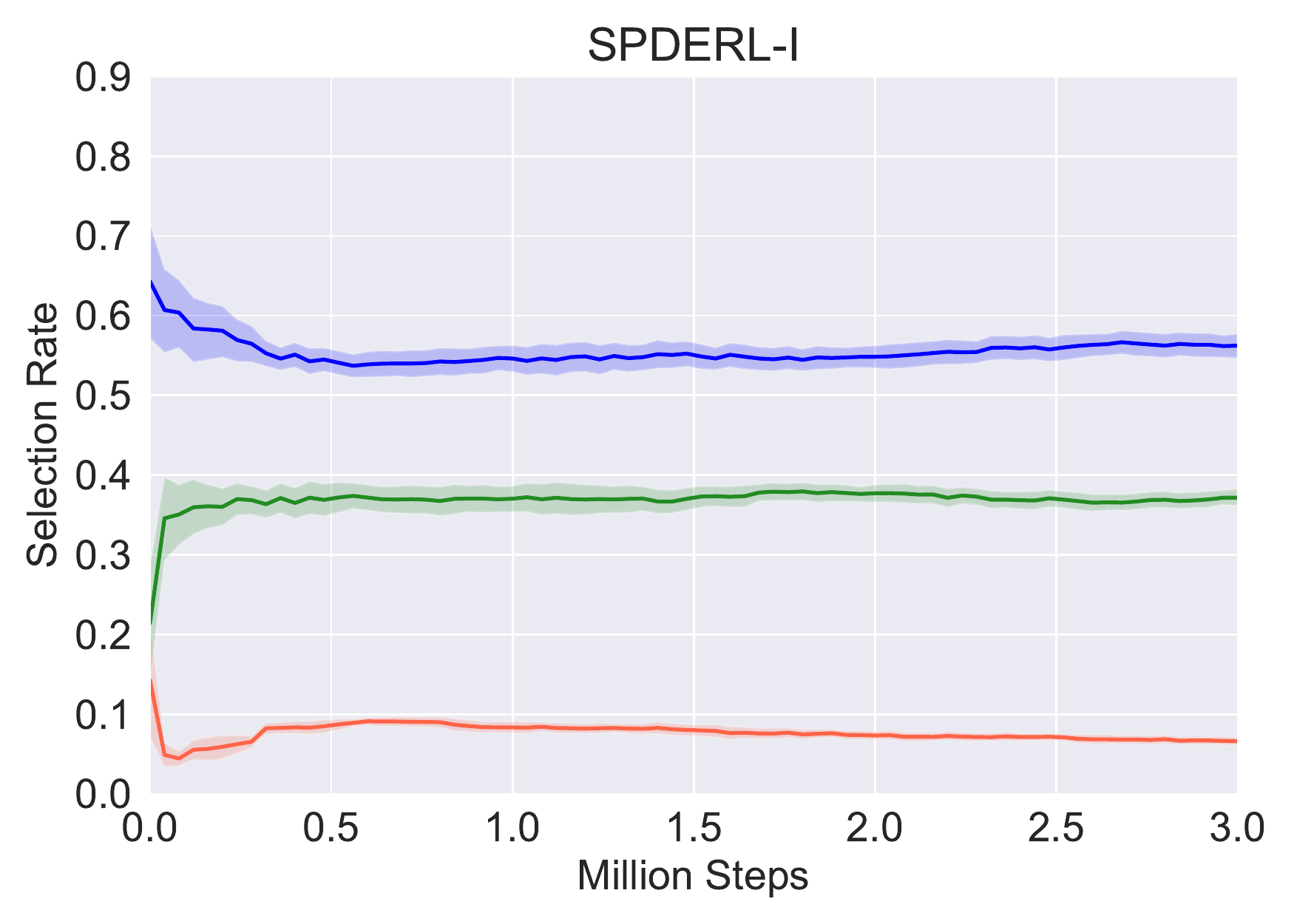} 
\caption{Accumulative rates of the RL agent being selected, discarded, or chosen as the elite during the training process in Hopper. The results indicate that SC can potentially stabilize the internal dynamics of the original frameworks.}
\label{fig7}
\end{figure*}

\begin{figure*}[h!]
\centering
\subfigure[HalfCheetach agent]{
\begin{minipage}[t]{0.99\textwidth}
\centering
\includegraphics[width=\textwidth]{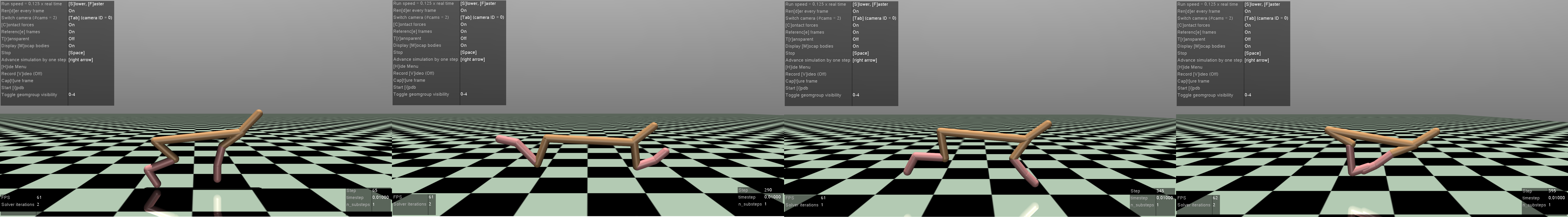}
\end{minipage}
}
\subfigure[Hopper agent]{
\begin{minipage}[t]{0.99\textwidth}
\centering
\includegraphics[width=\textwidth]{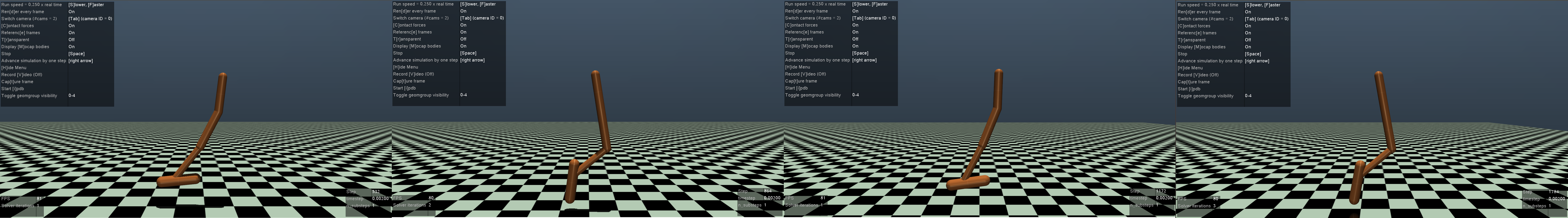}
\end{minipage}
}

\caption{Several intriguing behavioral patterns of the agents trained by SPDERL-I and SPDERL-G. (a) A HalfCheetah agent trained by SPDERL-I with average performance of 14000 points over 50 test seeds. The agent is able to adjust its posture more appropriately and run faster. (b) A Hopper agent trained by SPDERL-G with average performance of 4100 points over 50 test seeds. The agent jumps faster and learns to better stabilize the center of gravity.}
\label{behavior}
\end{figure*}

\textbf{Interactions between RL and EA.} To gain a deeper insight into the internal transformation of hybrid frameworks in the presence of SC, we highlight the changes in the internal dynamic between the RL agent and the genetic population. We keep a separate record of the accumulative rates of the RL actor being selected, discarded, or chosen as an elite in the population in Hopper. In Figure \ref{fig7}, although ERL, SERL-G, and SERL-I present similar dynamics, the integration of SC significantly stabilizes the internal dynamic of learning and evolution, which is more pronounced in PDERL and the two variants of SPDERL, where the evolutionary search is the major driving force for the training process. It is reasonable to hypothesize that optimization based both on the real and predicted fitness function may benefit the evolutionary search and make the optimization process more stable.

\textbf{Intriguing behavioral patterns.} A notable discovery is that the agents trained by our proposed methods can produce highly intriguing behavioral patterns than the original hybrid approaches, as shown in Figure \ref{behavior}. The solutions found by hybrid frameworks with SC are more intriguing and stable. On some control tasks, they can better adapt themselves to the environments and perform more competently. 

\begin{figure}
\centering
\includegraphics[width=0.49\columnwidth]{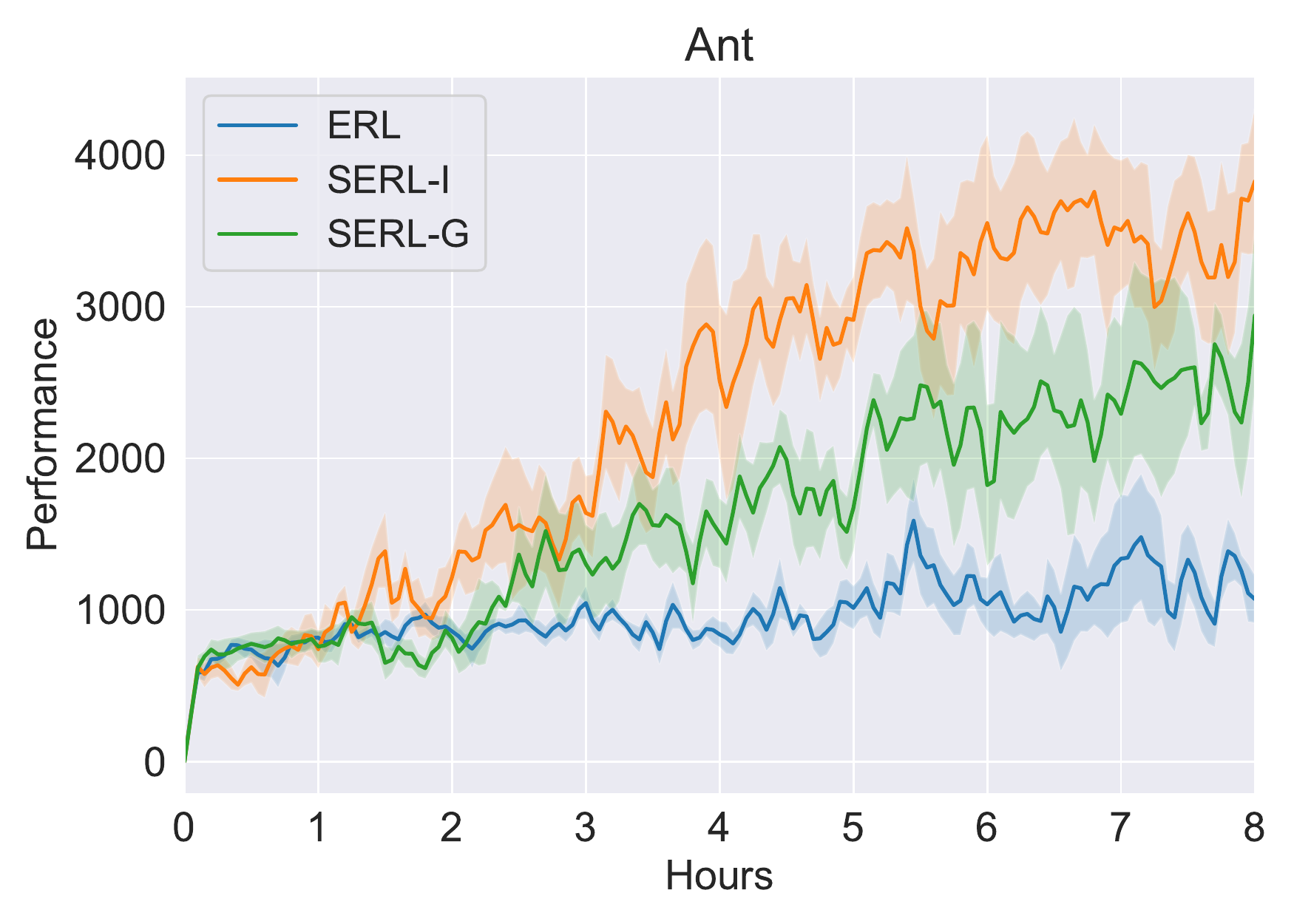}
\includegraphics[width=0.49\columnwidth]{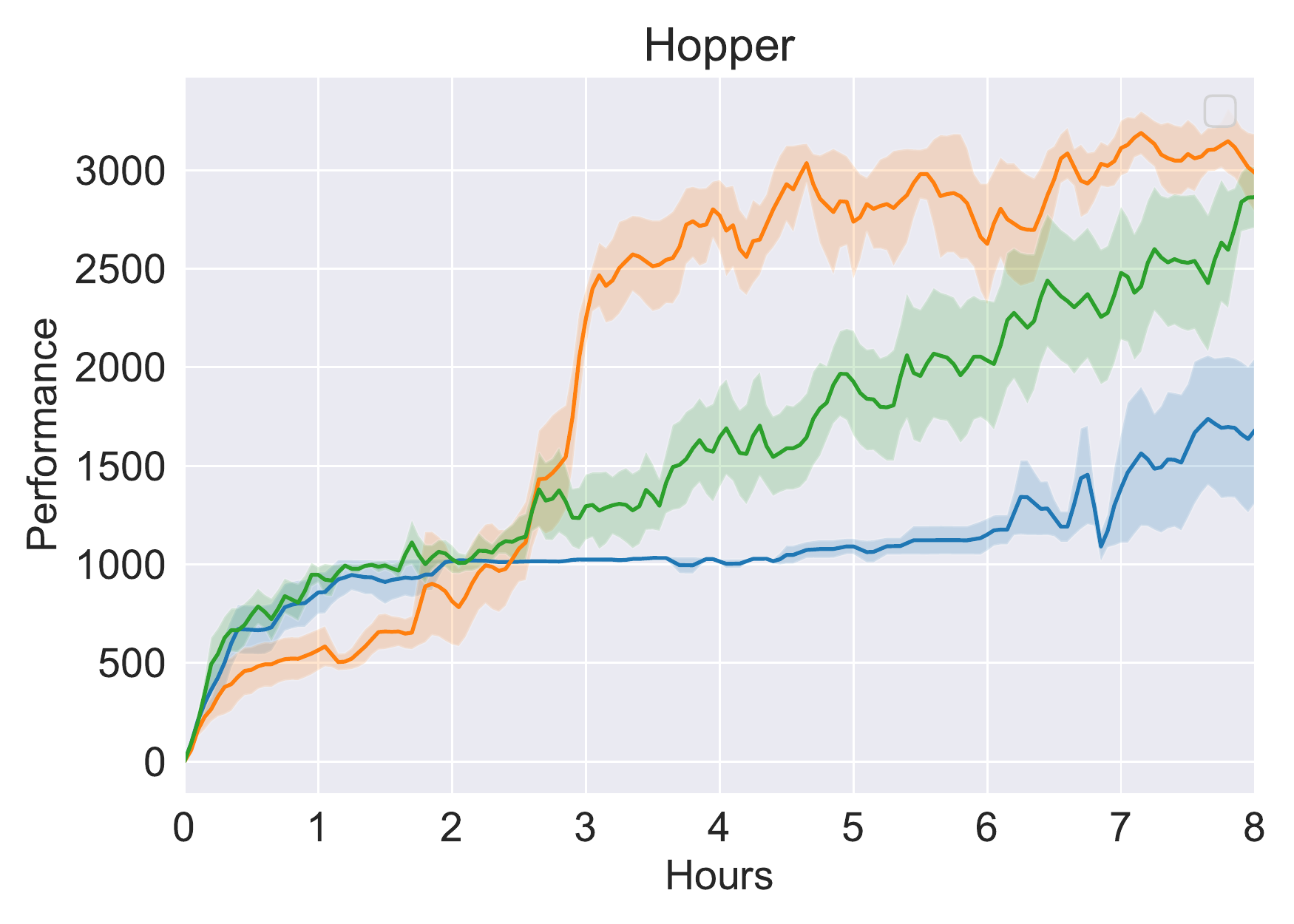}
\caption{Comparison of ERL, SERL-I and SERL-G in terms of the training time and performance in Ant and Hopper.}
\label{time}
\end{figure}

\begin{figure}
\centering
\includegraphics[width=\columnwidth]{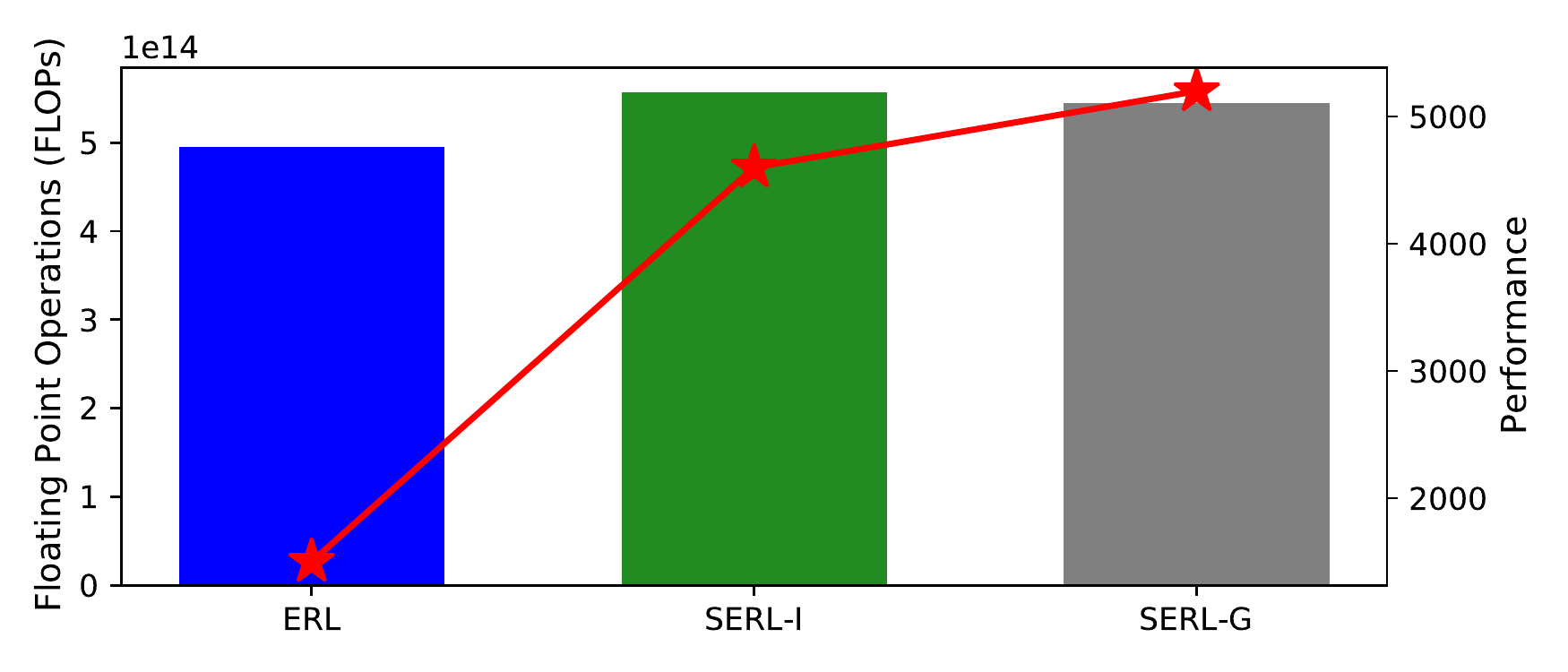}
\caption{Comparison of ERL, SERL-I and SERL-G in terms of the number of FLOPs and performance (red star) achieved at 3M environment steps in Walker.}
\label{flops}
\end{figure}

\subsection{Computational efficiency}
The number of real evaluations is potentially limited due to time and/or money, it is critical to make full use of available resources to achieve expected performance. In this part, we aim to validate the improvement of computational efficiency of original hybrid frameworks when SC is introduced, mainly from the following aspects: the sample reduction, time consumption and the Floating Point Operations (FLOPs).

\begin{table*}[t]
\caption{The average time steps (in million, equal to the number of consumed samples) of different algorithms in various environments when reaching the target score. ``$-$” represents that DDPG or GA can not reach the target.} 
\label{tab:sample}
\centering 
\fontsize{10}{9}\selectfont   
\begin{threeparttable} 
\resizebox{\textwidth}{!}{
\begin{tabular}{lc|cc|ccc|ccc}  
\toprule       
{\bf Env}&{\bf Score}&{\bf DDPG}&{\bf GA}&{\bf ERL}&{\bf SERL-G}&{\bf SERL-I}&{\bf PDERL}&{\bf SPDERL-G}&{\bf SPDERL-I}\cr
\midrule 
Ant         & $5,000$    &$-$ &$-$ & $4.266$& \bm{$1.564$} & $\bm{1.685}$ & $1.464$ & $1.784$& $ \bm{1.085}$\cr
Hopper      & $2,500$    &$-$ &$-$ & $2.519$& $\bm{1.073}$& $\bm{0.786}$ & $0.857$& $\bm{0.554}$& $\bm{0.561}$\cr
Walker      & $2,000$    &$0.775$ &$-$ & $3.572$& $\bm{0.767} $& $\bm{0.505}$ & $1.312 $& $\bm{0.796}$& $\bm{0.508}$\cr
Swimmer     & $300$      &$-$ &$2.640$ & $0.968$& $\bm{0.516}$& $1.034$ & $0.742$& $1.489$& $3.568$\cr
Reacher     & $-5$       &$-$ &$-$ & $1.051$& $\bm{0.322}$& $\bm{0.224}$ & $0.717 $& $ \bm{0.513}$& $ \bm{0.672}$\cr
HalfCheetah & $10,000$   &$0.986$ &$-$ & $4.092$& $\bm{2.565}$& $\bm{1.761}$ & $2.234$ & $\bm{2.153}$& $\bm{2.201}$\cr
\bottomrule 
\end{tabular}
}
\end{threeparttable}
 \end{table*}

\textbf{Sample consumption.} We conduct a sample reduction study of different methods in various environments when reaching the target scores. The number of environmental interactions (time steps) is equal to the number of consumed samples, and the results are reported over 6 suns. According to Table \ref{tab:sample}, SC can significantly reduce the sample consumption across most domains, especially in the environment with relatively higher noise and the reward variance like Walker and Hopper. For instance, ERL needs to perform almost 3 million environmental interactions more than SERL-G and SERL-I in Walker to reach the same score, and PDERL needs to consume 3 hundred thousand interactions more than SPDERL-G and SPDERL-I in Hopper. For different control strategies, the individual-based control requires a relatively small number of samples compared with generation-based control, due to its mechanism of pre-selecting potential high-performing individuals, which can be subsequently presented in real environments to generate higher-quality samples, as discussed in Section \ref{indie}. In a nutshell, SC can further improve the sample-efficiency of the original hybrid frameworks, as the diverse historical data is not only used for training the DRL algorithm but also employed to help evaluate the genetic population.   

\textbf{Time consumption.} We focus on the training time and corresponding performance of different methods, especially on the time costs reduced by the surrogate model. As SC is applied on top of the original algorithms, there are no changes in the network structures of actors and critics. The real evaluation of the genetic population is serial, and the results reported are averaged over 6 runs, and the total training time of each run is limited to 8 hours. The parameter settings are: $\omega=0.6$ for SERL-G and $\alpha=1.0$ for SERL-I. According to Figure \ref{time}, in the Ant environment, it is hard for ERL to reach 3000 points within 8 hours, while SERL-G takes approximately 8 hours and SERL-I only takes 4.5 hours. In the Hopper environment, it takes 8 hours for ERL to reach above 1500 points, 7 hours for SERL-G to reach 2500 points, while SERL-I is much more time-efficient, achieving the same performance as SERL-G in only 3 hours. When using real fitness evaluation, the actor only needs to perform forward propagation through its policy neural network, such as the calculation of $\mu(s_{j}|\theta^{\mu})$, and the reward is provided by the environment itself, which may be time-consuming and expensive. While using the surrogate-assisted evaluation, additional forward propagation will be introduced to calculate the averaged value of $Q(s_{j},\mu(s_{j}|\theta^{\mu})|\theta^{Q})$ via diverse historical state information. SC can efficiently transfer the computational burden of real evaluations to additional forward propagation through the surrogate model, and the computational time of these calculations can be easily reduced by parallelization and shared memory.

\textbf{Floating Point Operations.} The computational efficiency of SC is finally verified by comparing the FLOPs consumed by ERL, SERL-I, SERL-G and their corresponding performance (Figure \ref{flops}). The neural networks in our work are all fully connected and the number of updates of the RL agent is equal to the environment steps. Thus, regardless of which hybrid framework that SC is combined with, the computational cost of the entire training process can always be divided into the surrogate-assisted evaluation cost and the original optimization cost, as shown in Appendix B. As mentioned above, while bringing significant performance improvement, SC is computationally efficient in that only a small amount of forward propagation for surrogate-assisted evaluations is introduced. 

\section{Conclusion and Future Work}\label{conclusion}
The application of hybrid RL frameworks to expensive learning problems has largely been limited by the cost of evaluating the population in real environments. In this work, we propose a surrogate-assisted controller with two management strategies, which can be easily integrated into existing hybrid frameworks to simultaneously facilitate the optimization of the RL agent and the evaluation of the population. To the best of our knowledge, this is the first attempt on introducing the surrogate model into hybrid RL frameworks. Empirical evaluations show that the combination of SC with two state-of-the-art evolutionary reinforcement learning frameworks ERL and PDERL can effectively reduce the evaluation cost and boost the performance. Furthermore, SC brings beneficial changes to the internal dynamics of learning and evolution, resulting in more collaborative interactions between the RL agent and the EA population.

Learning and evolution are two symbiotic counterparts in nature. It is of great scientific importance to fully appreciate and explore this hybrid paradigm towards implementing truly competent artificial intelligence. As to future work, there are plenty of fascinating directions to advance our proposed techniques, including real-world settings, such as embodied AI, designing self-adaptive evolution control strategies and more effective evaluation criteria for the surrogate. For complex and challenging settings, multi-agent evolutionary reinforcement learning with multiple surrogates is expected to further extend the horizon.

\section*{Appendix A. The Swimmer environment} 
\begin{figure}[t]
\centering
\includegraphics[width=\columnwidth]{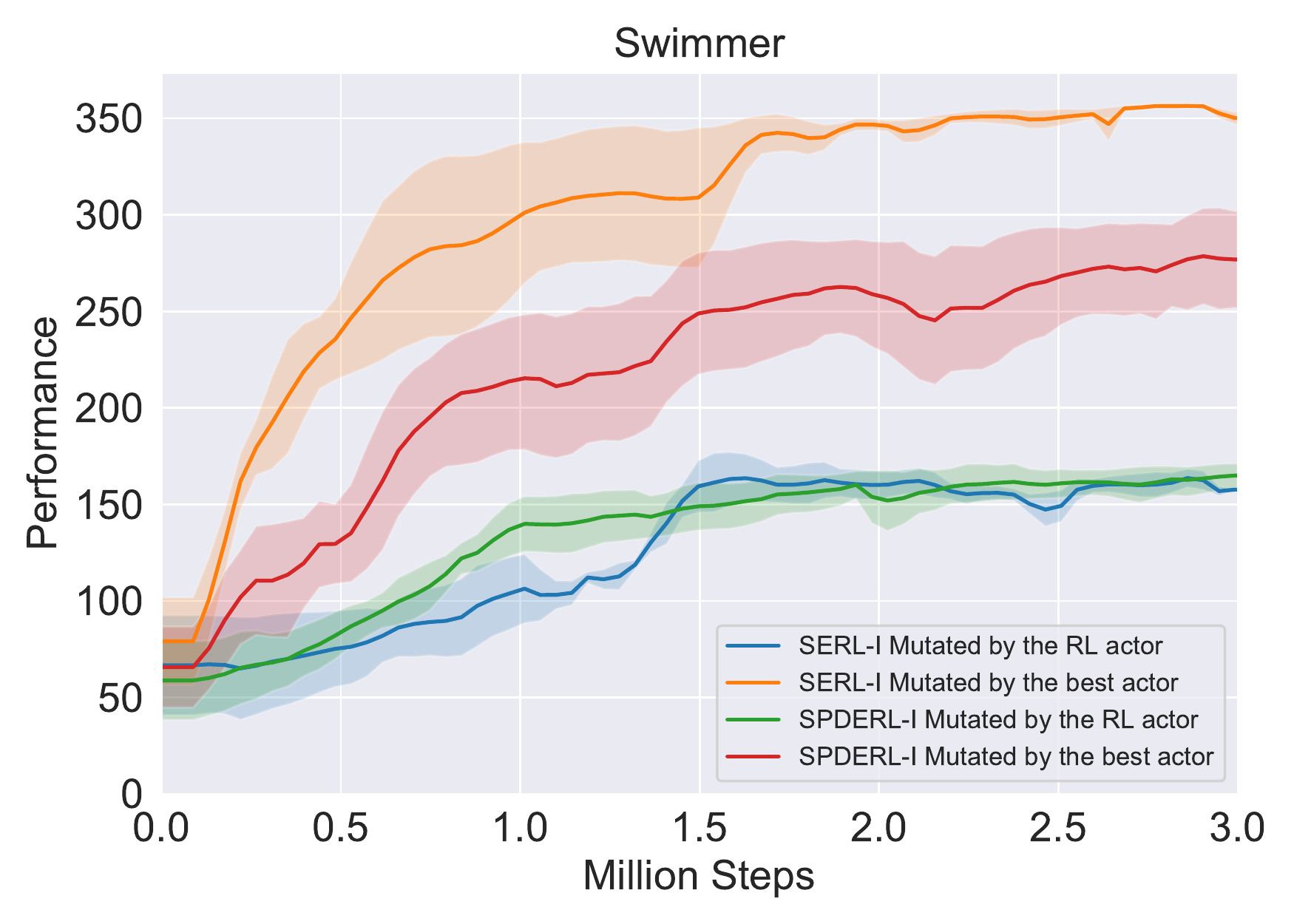} 
\caption{Learning curves using different approaches of generating the candidate population in the Swimmer environment.}
\label{swim}
\end{figure}

Results in Figure \ref{fitness} demonstrate that all individual-based control methods fail in the Swimmer environment because the RL agent is unfortunately misled by the deceptive gradient information, and candidates mutated from the RL actor could not provide useful information for policy improvement. To alleviate this issue, we generate the candidate population by mutating the best actor that has been found by genetic operations (Figure \ref{swim}). In this specific environment, the evolutionary search is more suitable for driving the optimization process.

\section*{Appendix B. Measurement of FLOPs} 
In this section, we introduce how the FLOPs values in Figure \ref{flops} are calculated. The neural networks in our work are all fully connected and the number of updates of the RL agent is equal to the environment steps. Similar to \cite{seo2021state}, we consider the FLOPs of each forward pass being half of that of the backward pass. We also assume that the consumed FLOPs of activation functions and operations that do not require passing through neural networks are negligible.
\begin{table}[t] 
\centering 
\caption{Symbolic meanings and values in the Calculation of FLOPs} 
\label{tab:flops}
\fontsize{9}{9}\selectfont   
\begin{threeparttable} 

\resizebox{\columnwidth}{!}{
\begin{tabular}{lc} 
\toprule         
{\bf Symbol}&{\bf Value}\cr
\midrule 
Environment steps $T$ & $3,000,000$\cr
Population size $n$ & $10$\cr
Candidate population size $n_{c}$ & $20$\cr
RL Agent batch size $b$ & $128$\cr
Evaluation memory size $k$ & $50,000$\cr
Evolutionary generations $G$ &  $240$\cr
Evolutionary generations using the surrogate $G_{s}$ & $390$\cr
Flops of forward propagation of RL-Actor $A_{f}$ & $11,136$\cr
Flops of forward propagation of RL-Critic $C_{f}$ & $249,800$ \cr
Flops of backward propagation of RL-Actor $A_{b}$ & $22,272$\cr
Flops of backward propagation of RL-Critic $C_{b}$ & $499,600$\cr
\bottomrule 
\end{tabular}}
\end{threeparttable}
\end{table}
We follow the procedures in \cite{molchanov2016pruning} to measure the FLOPs consumed by each method within 3M environment steps. The calculation formulas for FLOPs consumed by ERL, SERL-I and SERL-G are shown in Eqs.(\ref{eq:flops_erl}), (\ref{eq:flops_serli}) and (\ref{eq:flops_serlg}), respectively. 

\begin{equation}
F_{ERL}=T(A_{f}+b\times(2A_{f}+3C_{f}+C_{b}+A_{b}))
\label{eq:flops_erl}
\end{equation}

\begin{equation}
F_{SERL(I)}=F_{ERL}+G_{s}\times n\times k\times (A_{f}+C_{f})
\label{eq:flops_serli}
\end{equation}

\begin{equation}
F_{SERL(G)}=F_{ERL}+G\times n_{c}\times k\times (A_{f}+C_{f})
\label{eq:flops_serlg}
\end{equation}

All the symbols and their symbolic meanings that are involved in calculating are shown in Table \ref{tab:flops}.

\section*{Appendix C. Combining SC with CEM-RL}
\begin{figure*}[t]
\centering
\includegraphics[width=0.325\textwidth]{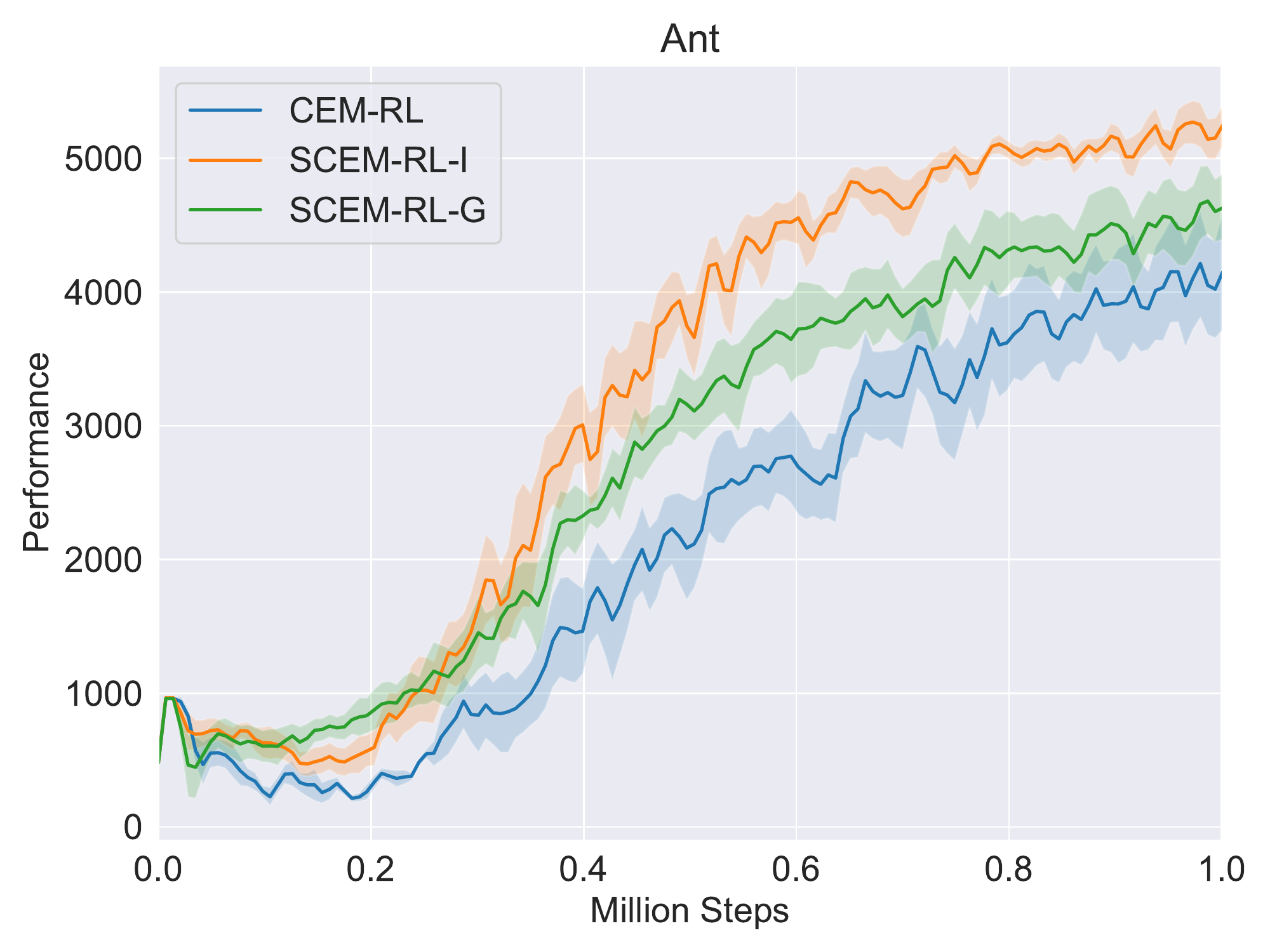}
\includegraphics[width=0.325\textwidth]{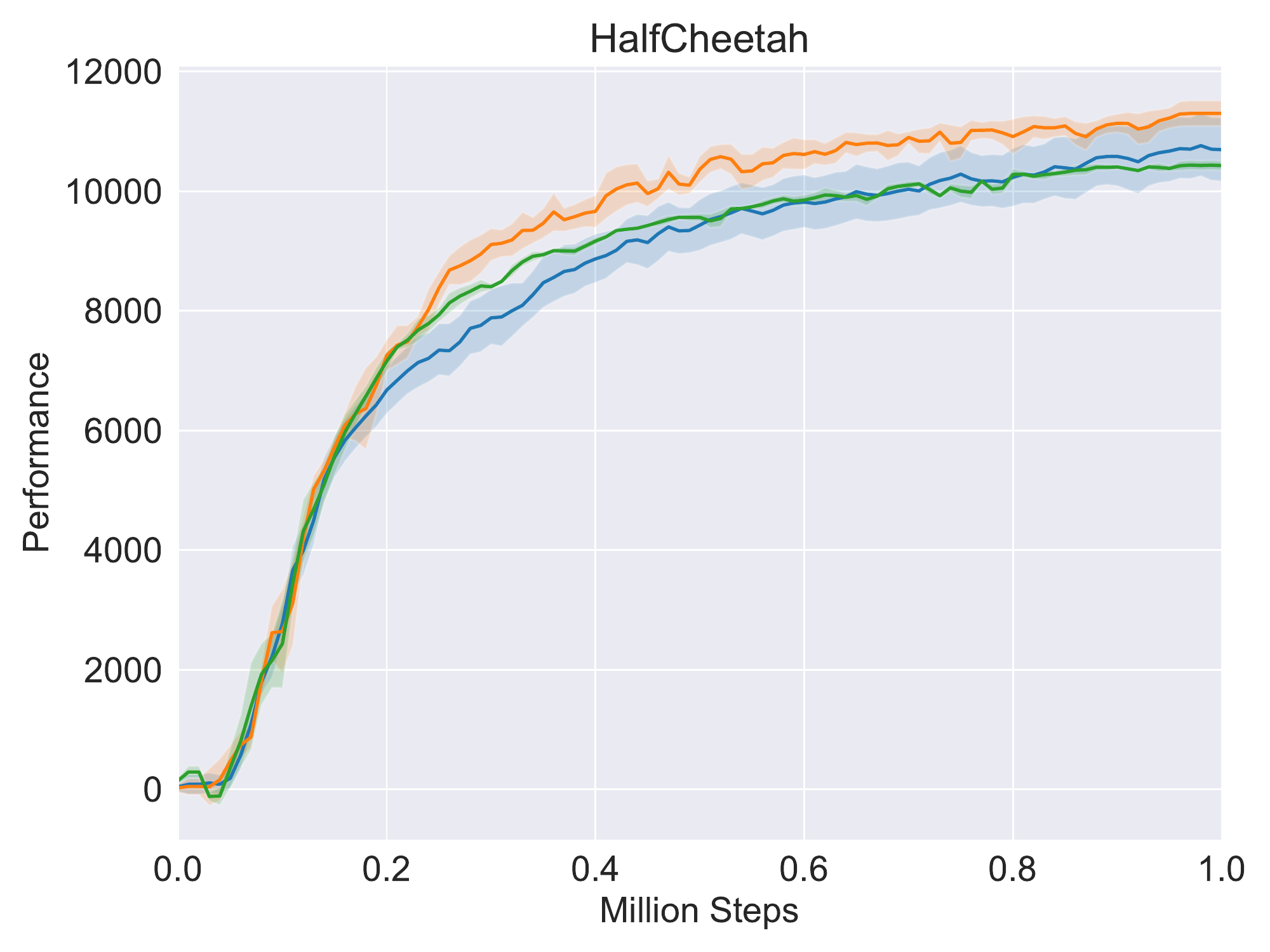} 
\includegraphics[width=0.325\textwidth]{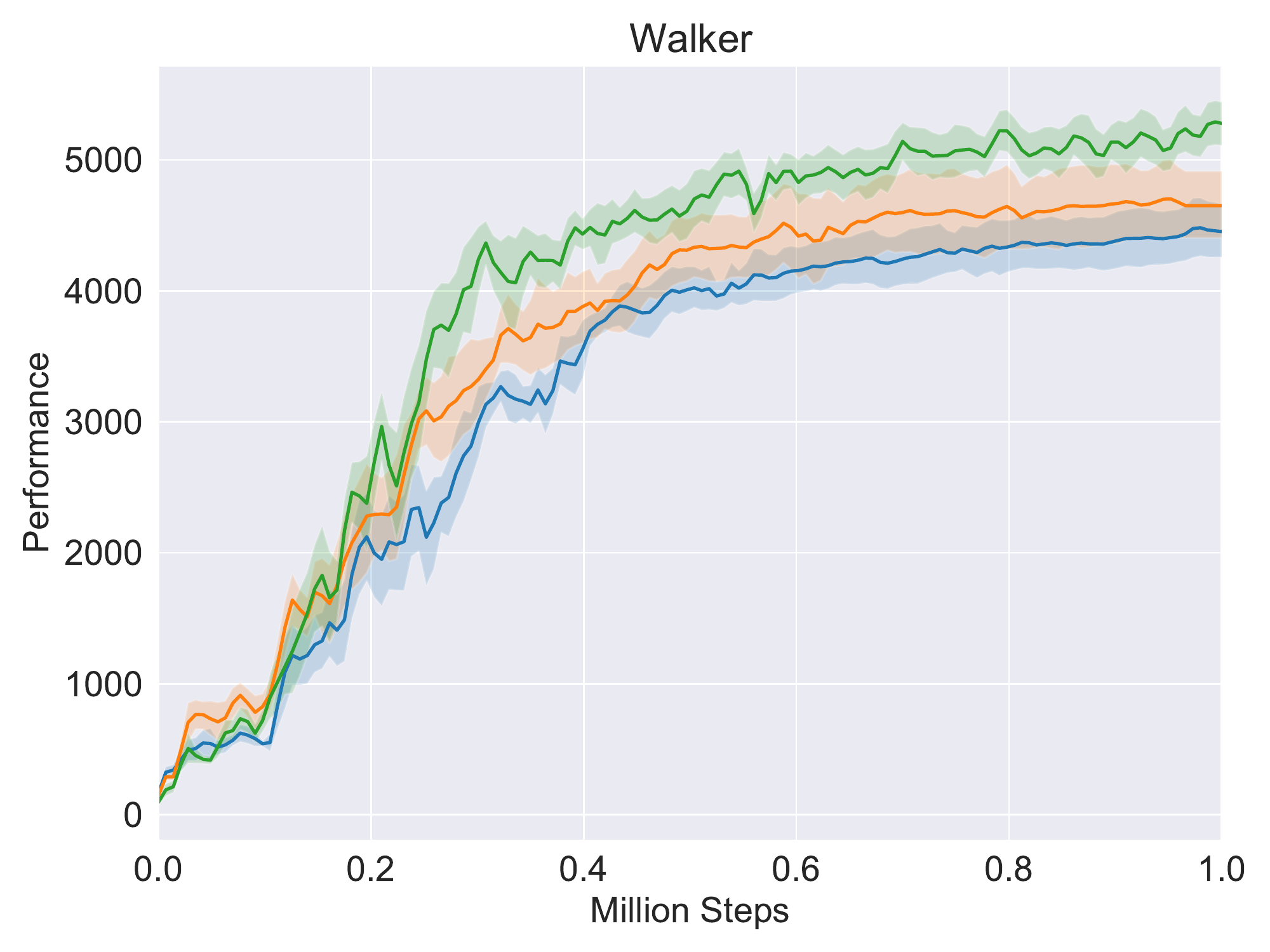}
\caption{Learning curves on three MuJoCo environments: Ant, HalfCheetah and Walker.}
\label{cemrl}
\end{figure*}

We conduct an additional experiment on combining SC with another state-of-the-art hybrid framework CEM-RL \cite{pourchot2018cem}, referred to as Surrogate-assisted CEM-RL (SCEM-RL), to show SC can be applied to a different type of EA. The EA parts of ERL and PDERL are based on GA, while in CEM-RL, it is based on Cross-Entropy Method (CEM), which is a typical Evolutionary Strategy (ES). CEM-RL combines CEM with TD3 learners \cite{fujimoto2018addressing}, an off-policy DRL algorithm related to DDPG \cite{lillicrap2015continuous}. 

We train SCEM-RL-G (SCEM-RL with generation-based control) and SCEM-RL-I (SCEM-RL with individual-based control) with 6 different random seeds. Figure \ref{cemrl} illustrates their learning curves during training on three control tasks from MuJoCo \cite{todorov2012mujoco}. Apart from the clear advantages of SC in the training performance and sample consumption, similar to the results in Section \ref{overall}, on HalfCheetah and Ant, the improvement over the original hybrid framework is more evident under the individual-based control, while the generation-based control method is more favorable on Walker.

\printcredits

\bibliographystyle{plain}
\bibliographystyle{elsarticle-num.bst}

\bibliography{cas-refs.bib}

\end{document}